\documentclass[final,3p,12pt]{elsarticle}
\journal{Computer Methods in Applied Mechanics and Engineering}

\let\today\relax
\makeatletter
\def\ps@pprintTitle{%
    \let\@oddhead\@empty
    \let\@evenhead\@empty
    \def\@oddfoot{\footnotesize\itshape
         {Published in Computer Methods in Applied Mechanics and Engineering} \hfill\today}%
    \let\@evenfoot\@oddfoot
    }
\makeatother

\usepackage[utf8]{inputenc}
\usepackage{textcomp}
\usepackage{hyperref}
\usepackage{graphicx}
\graphicspath{ {figures/} }
\usepackage{amsmath}
\usepackage{bm}
\usepackage[version=4]{mhchem}
\usepackage{booktabs}
\usepackage{float}
\usepackage{siunitx}
\usepackage{algpseudocode}
\usepackage{algorithm}
\usepackage{enumitem}
\usepackage{optidef}
\usepackage{color}
\usepackage{xcolor}
\usepackage{longtable,tabularx}
\usepackage{subfig}
\usepackage{lineno}
\graphicspath{{figures}/}
\usepackage[ruled,vlined,algo2e]{algorithm2e}

\usepackage{amssymb}

\begin{document}

\begin{frontmatter}



\title{RiemannONets: Interpretable Neural Operators for Riemann Problems}

\author{Ahmad Peyvan}
\author{Vivek Oommen}
\author{Ameya D. Jagtap}

\author{George Em Karniadakis}
\affiliation{organization={Division of Applied Mathematics, 182 George Street, Brown University},
    city={Providence},state={RI}, postcode={02912},country={USA}
            }

\begin{abstract}
Developing the proper representations for simulating high-speed flows with strong shock waves, rarefactions, and contact discontinuities has been a long-standing question in numerical analysis.
Herein, we employ neural operators to solve Riemann problems encountered in compressible flows for extreme pressure jumps (up to $10^{10}$ pressure ratio). In particular, we first consider the DeepONet that we train in a two-stage process, following the recent work of \cite{lee2023training}, wherein the first stage, a basis is extracted from the trunk net, which is orthonormalized and subsequently is used in the second stage in training the branch net. This simple modification of DeepONet has a profound effect on its accuracy, efficiency, and robustness and leads to very accurate solutions to Riemann problems compared to the vanilla version. It also enables us to interpret the results physically as the hierarchical data-driven produced basis reflects all the flow features that would otherwise be introduced using ad hoc feature expansion layers. We also compare the results with another neural operator based on the U-Net for low, intermediate, and very high-pressure ratios that are very accurate for Riemann problems, especially for large pressure ratios, due to their multiscale nature but computationally more expensive. 
Overall, our study demonstrates that simple neural network architectures, if properly pre-trained, can achieve very accurate solutions of Riemann problems for real-time forecasting. The source code, along with its corresponding data, can be found at the following URL: \url{https://github.com/apey236/RiemannONet/tree/main}

\end{abstract}




\end{frontmatter}

{\renewcommand\arraystretch{0.75}

\section{Introduction}

In recent years, data-driven modeling methods have shown great potential for solving many challenging problems in the fields of computational science and engineering. Some of these methods, in particular, can have a large impact on large-scale computational problems and real-time forecasting as they are efficient and, once trained, can be used for the same or similar tasks repeatedly. Neural operators are a new paradigm for learning nonlinear mappings between the input and output functions. In particular, there are various neural operators available in the literature: The \textbf{Deep} \textbf{O}perator \textbf{Net}work (DeepONet), developed by Lu et al.  \cite{lu2021learning} (first published in 2019 in \cite{lu2019deeponet}), is the first proposed neural operator. DeepONet has been used to solve many problems in computational science and engineering, such as stiff chemical kinetics \cite{goswami2024learning}, multiscale bubble dynamics \cite{lin2021operator}, brittle fracture analysis \cite{goswami2022physics}, two-phase microstructural evolution \cite{oommen2022learning} solar-thermal systems forecasting \cite{osorio2022forecasting}, electroconvection \cite{cai2021deepm}, etc. In addition, several extensions of DeepONet have been proposed in recent studies, including Partition-of-Unity (PoU) based DeepONet \cite{goswami2024learning}, physics-informed DeepONet \cite{wang2021learning}, DeepONet with proper orthogonal decomposition (POD-DeepONet) \cite{lu2022comprehensive, venturi2023svd}, multifidelity DeepONet \cite{howard2022multifidelity,lu2022multifidelity}, DeepONet with UQ \cite{yang2022scalable,moya2023deeponet,lin2023b} multiscale DeepONet \cite{liu2021multiscale}, etc. Other neural operator networks have also been proposed in the literature, such as the Fourier neural operator (FNO) \cite{li2020fourier}, the wavelet neural operator (WNO) \cite{tripura2022wavelet}, the spectral neural operator (SNO) \cite{fanaskov2022spectral}, the convolutional neural operator \cite{raonic2023convolutional}, etc. The main difference between DeepONet and the aforementioned operators is that DeepONet learns a new basis (through the trunk net)  to represent the operator whereas other operators, e.g., FNO, use a pre-specified basis, e.g., Fourier expansions.

When implemented on a computer, not all models behave as operators, raising doubts about what operator learning actually is. To address this, \cite{bartolucci2023neural} proposed a unifying mathematical framework for representation equivalent neural operators (ReNO) to ensure that operations at the continuous and discrete levels are equivalent. In their recent work, Lee and Shin \cite{lee2023training} proposed a novel two-step training procedure for DeepONet. The newly introduced sequential two-step training approach begins with trunk network training, which includes Gram-Schmidt orthonormalization via QR-factorization, and then advances on to branch network training.

U-Net-based operators \cite{gupta2022towards, oommen2023rethinking, ovadia2023ditto, rahman2022u} are another class of neural operators that can be particularly effective for approximating mathematical operators due to their inherent multi-scale nature. U-Net is a U-shaped fully convolutional neural network \cite{ronneberger2015u}, first proposed for biomedical image segmentation tasks. The authors of \cite{gupta2022towards, oommen2023rethinking, ovadia2023ditto} demonstrated that conditioning the U-Net with respect to time can significantly improve the predictions of the vanilla U-Net. In our current work, we extend this idea and condition the U-Net with respect to the initial pressure and temperature states.

Neural operators have been successfully used to solve high-speed viscous flows. In \cite{mao2021deepm}, Mao et al. employed DeepONet to solve hypersonic viscous flows that give high-gradient solutions without exact shocks.
In this work, we are using DeepONet to solve the Riemann problem for the compressible Euler equations of gas dynamics. To the best of our knowledge, this is the first attempt to solve the Riemann problem that has discontinuous solutions using neural operator networks.
Specifically, the Riemann problem presents a hyperbolic partial differential equation characterized by a discontinuous initial solution. The inherent complexity of dealing with discontinuities makes these problems particularly challenging to solve. Notably, the accuracy of the solution tends to deteriorate in the vicinity of shocks and contact waves.

In summary, the following are our main contributions:
\begin{itemize}
    \item Our study leverages the capabilities of deep neural operators to investigate its efficacy in mapping input pressure ratios to the final solution at a specified time. The final solution encompasses primitive field variables such as density, pressure, and velocity.
    \item We assess the impact of activation functions on prediction accuracy by investigating both fixed and adaptive (Rowdy) activation functions \cite{jagtap2022deep}.
    \item We explore the performance of two training strategies for RiemannONet: the traditional vanilla approach (one-step training) and the recently proposed novel two-step training method \cite{lee2023training}. 
    \item We enforce positivity-preserving constraints during the training of RiemannONet, a crucial consideration grounded in the governing physical principles.
    \item We systematically compare two types of neural operators for solving Riemann problems: 1) a modified DeepONet architecture, and 2) a U-Net conditioned on the pressure and temperature initial conditions.
    \item We obtain interpretable basis functions for such discontinuous solutions. To this end, we employ QR and SVD methods to investigate the solution spectrum and diverse bases. Furthermore, we conduct a comprehensive investigation into these basis functions, delving into the influence of network architecture on their structure.
\end{itemize}

The structure of this paper unfolds as follows: Section 2 elucidates the governing equations for the Riemann problem. In Section 3, we describe the two neural operators employed in RiemannONets. Section 4 is dedicated to an in-depth discussion of the results obtained across various test cases. Section 5 delves into the exploration of hierarchical and interpretable basis functions for representing the two neural operators. Our conclusions and a summary of our findings are encapsulated in Section 6.

\section{Governing Equations}
In this study, we consider the Riemann problem of the one-dimensional general form of the hyperbolic Euler equations \eqref{eq:general} as

\begin{equation}
\frac{\partial \mathbf{U}(x,t)}{\partial t}+\frac{\partial \mathbf{F}(\mathbf{U}) }{\partial x}=\mathbf{S}(x,t),\quad t\in[0,t_f],\quad \textrm{and}\quad \mathbf{U}(x,0)=\Bigg\{\begin{array}{lc}
        \mathbf{U}_L & x\le x_c \\
        \mathbf{U}_R & x > x_c
    \end{array}
\label{eq:general}
\end{equation}
where $t$ is time, $x$ is the spatial coordinate, $\mathbf{U}$ indicates the vector of conservative variables, $\mathbf{F}$ denotes the corresponding advective fluxes, and $\mathbf{S}$ is a vector of source terms. The hyperbolic Euler equation is subjected to a discontinuous initial condition composed of two constant states of the conservative variable vectors as $\mathbf{U}_L$ and $\mathbf{U}_R$ separated at $x=x_c$. We are interested in determining the solution at a final time, as $t=t_f$. The definitions of the conservative variable, flux, and source term vectors are explained below. The Euler equations takes the form of Eq.~\eqref{eq:general} with $\mathbf{S}=\mathbf{0}$, the conservative variables vector and flux vectors defined as 

\begin{equation}
\mathbf{U}=\begin{pmatrix}\rho\\ \rho u \\ \rho E
\end{pmatrix}^T, \quad \mathbf{F}=\begin{pmatrix}\rho u\\ \rho u^2+p \\ u(\rho E+p)
\end{pmatrix}^T,
\label{eq:vec}
\end{equation}
respectively. In Eq.~\eqref{eq:vec}, the total energy is described as 

\begin{equation}
\rho E =\frac{p}{\gamma-1}+\frac{1}{2}u^2,
\label{eq:enochem}
\end{equation}
where $\gamma=1.4$ for all the test cases without chemistry. In Eqs.~\eqref{eq:vec} and \eqref{eq:enochem}, $\rho$ is density, $p$ is pressure, and $u$ indicates the velocity.

\newcommand{\ocircleplus}{\mathrel{\ooalign{$\bigcirc$\cr\hidewidth$+$\hidewidth\cr}}}

\section{Methodology}
Within this section, we introduce two distinct neural operators: DeepONet and U-Net. We provide a comprehensive overview of the necessary modifications essential for their effective training within the framework of RiemannONets.

\subsection{Deep Operator Networks}
The development of modern machine learning models has made it possible to create rapid simulators for solving parametric PDEs. 
Instead of dealing with the PDE explicitly, a substitute model of the PDE solution operator that can regularly simulate PDE solutions for varied initial and boundary conditions is frequently required, e.g. in inverse problems, in design, in uncertainty quantification, etc. This promise is fulfilled by the novel idea of neural operators, which was first proposed in 2019 in \cite{lu2019deeponet} in the form of DeepONet \cite{lu2021learning} and is inspired by the universal approximation theorem of operators. In this section, we will discuss the architecture of DeepONet for the Riemann problem named RiemannONet.

\begin{figure}[htpb]
  \centering
     \includegraphics[clip, scale=0.47]{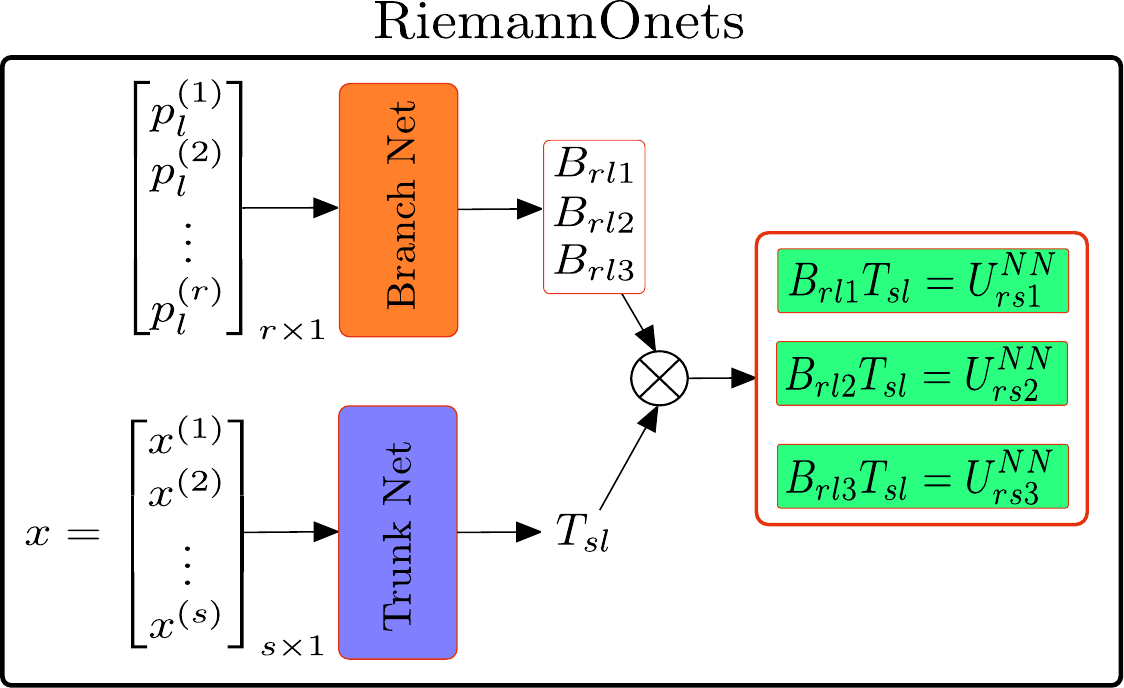}
     \caption{Schematic representation of RiemannONet for the Sod's problem. The input to the trunk net is the sensor location in spatial dimensions, while the input to the branch net is different realizations of left side pressure (keeping right side pressure fixed). The output of RiemannONet is the primitive variables consisting of velocity, density, and pressure at the final time.}
     \label{fig:RON}
\end{figure}

\subsubsection{Training procedure}

\paragraph{Single step}
Our RiemannONet consists of two networks, namely, the trunk net and the branch net; see Fig. \ref{fig:RON}. In the branch net, the encoded input is the different realization of input pressure $p_l$ (only the left side, as the right side has a fixed value), whereas the trunk net input consists of the spatial dimensions $x$. The aim is to learn the solution (primitive variables) at a later time. The output of the network is given by

\begin{equation}
U_{rsj}^{NN}(\beta,\Theta) = B_{rlj}(\beta) T_{sl}(\Theta), ~~~ j = 1,2,3,
   \label{eq:deep} 
\end{equation}
where $U_{rsj}^{NN}$ is the profile of the $j^{th}$ primitive variable, $B_{rlj}$ is the output of the branch net, and $T_{sl}$ is the output of the trunk net. Here $l$ is the index of the latent dimension, which is a hyperparameter. In Eq.~\eqref{eq:deep}, subscript $r$ indicates the $r^\textrm{th}$ realization of the training dataset and $s^\textrm{th}$ coordinate point of the trunk net input. The symbol $\beta$ indicates all the trainable parameters of the branch network while $\Theta$ denotes trainable parameters of the trunk network.
The RiemannONet is being trained on a data set consisting of input-output data, which is divided into training and testing datasets. The loss function consists of a data mismatch term:

\begin{equation}
    \mathcal{J}(\beta,\Theta) = \left|\left|U_{rsj}^{\text{Exact}}-U_{rsj}^{NN}(\beta,\Theta)\right|\right|_{2},
\end{equation}
where $U_{rsj}^\textrm{Exact}$ is ground truth solution tensor containing all primitive variables.


\paragraph{Two-step}
The training of DeepOnet can be split into two stages employing the approach of Lee and Shin \cite{lee2023training}. First, the trunk net is trained by minimizing the loss function
\begin{equation}
    \mathcal{L}^t(\Theta,\mathbf{A}) = \left|\left|T_{sl}(\Theta)A_{rlj}-U_{rsj}^{\text{Exact}} \right|\right|_{2},
    \label{eq:firststep}
\end{equation}
where $\Theta$ represents all the trainable parameters of the trunk network and $\mathbf{A}$ denotes the elements of a trainable matrix defined as $\mathbf{A}\in \mathbb{R}^{r\times K\times v}$ with $K$ the number of neurons of the output layer of the trunk net and $v$ the number of all the primitive variables.
In the equation \eqref{eq:firststep}, the index $l$ iterates over the neurons in the final layer of the trunk network. Let $\Theta^*$ and $\mathbf{A}^*$ be the optimal values of the trainable parameters. The second step consists of training the branch net by first performing a  $QR$-factorization of the trunk net represented by the matrix $\mathbf{T}(\Theta^*)\in \mathbb{R}^{s\times K}$ using the following formula:
\begin{equation}
\mathbf{Q}^*\mathbf{R}^* = \mathbf{qr}\left(\mathbf{T}(\Theta^*)\right),
    \label{eq:qreq}
\end{equation}
where $\mathbf{qr}$ represents the QR-factorization function. In this study, we replaced the QR-factorization with the SVD decomposition since the SVD approach provides a unique solution and generates a hierarchical set of orthonormal basis functions. We can construct the equivalent Q, and R matrices for the SVD method as 
\begin{equation}
\mathbf{U}\mathbf{S}\mathbf{V} = \mathbf{svd}\left(\mathbf{T}(\Theta^*)\right),\quad \mathbf{Q}^*= \mathbf{U}, \textrm{ and} \quad \mathbf{R}^*=\mathbf{S}\mathbf{V}.
    \label{eq:svdeq}
\end{equation}
Therefore, we can replace the QR with the SVD decomposition without any other modification. The column of matrix $\mathbf{Q}^*$ forms a set of orthonormal basis functions.

Next, the branch net is trained to match $U^B_{rij}=R^*_{il}A^*_{rlj}$ where subscript $i$ and $l$ ranges are the same and equal to the number of neurons of the last latent layer. The tensor multiplication occurs along the second axis with a dimension of $K$. Hence, we minimize the loss function for the training of the branch net as

\begin{equation}
    \mathcal{L}^b(\mu) = \left|\left|B_{rlj}(\mu)-U_{rlj}^B\right|\right|_{2},
    \label{eq:branchloss}
\end{equation}
where $\mathbf{B}$ represents the branch network with $\mathbf{B}\in \mathbb{R}^{r\times K\times v}$. We can now construct a trained DeepONet model by multiplying a trunk network defined as $\mathbf{\Hat{T}}=\mathbf{T}(\Theta^*)\left(\mathbf{R}^{*}\right)^{-1}$ into the trained branch network defined as $\mathbf{B}(\mu^*)$. The trained branch network is also described as $\mathbf{B}(\mu^*)$, where $\mu^*$ is the optimum value of the branch network parameters that minimize the loss function in Eq. \eqref{eq:branchloss}. Algorithm \ref{alg:2step} gives the pseudo-code for two-step training procedure for DeepONet.

\begin{algorithm}[H]
\caption{Two-step training procedure for DeepONet}\label{alg:2step}
\SetAlgoLined
 \textbf{Given}: $p_l$ and vector of x-coordinates.  \\
\textbf{Step 1}: Train the trunk network using $\mathbf{A} \in\mathbb{R}^{r\times K \times v}$ and match the ground truth as Eq.~\eqref{eq:firststep}.  \\
 \textbf{Step 2}: Perform QR or SVD on the trained trunk network. \\
  \textbf{Step 3}: Compute $U^B_{rij}=R^*_{il}A^*_{rlj}$. \\
   \textbf{Step 4}: Train the branch network to match $U^B_{rij}$ as in Eq.~\eqref{eq:branchloss}.\\
\textbf{Output}: Use the trained trunk and branch networks to infer the density, velocity, and pressure profiles. \\

\end{algorithm}

\subsubsection{Rowdy activation functions}
Adaptive activation functions are state-of-the-art activation functions that can give superior performance with respect to their fixed counterparts \cite{raissi2019physics}. There are various adaptive activation functions proposed for deep and physics-informed neural networks \cite{jagtap2023important}; see, for example, the work of Jagtap et al. \cite{jagtap2020adaptive,jagtap2020locally}.
In this work, we employed the \textit{Rowdy} activation function \cite{jagtap2022deep}. These adaptive activation functions have been successfully employed to solve problems involving high-frequency complex structures. In the Rowdy activation function, the base activation function $\phi_1$ is any standard activation function, such as hyperbolic tangent. The rest of the activation functions $\phi_k, k = 2,\cdots, K$ are defined as sine functions as follows:
$$ \phi_k(x) = n \sin((k-1)nx), $$
where $n$ is the scaling factor, and here we select $n=10$. In this work, we modify the Rowdy activation function by adding a shiftable parameter to the base and sine functions. We choose $K=2$ and define the activation function as 
\begin{equation}
g(L)= h{\left(10a\left[L\right]+c\right)}+10a_1\sin{\left(10F_1\left[L\right]+c_1\right)},
    \label{eq:activation}
\end{equation}
where $L:=wx+b$ is the input to the activation function and $h(10a\left[L\right]+c)$ denotes the base function of the Rowdy activation functions that could be $\cos(10a\left[L\right]+c)$ or $\tanh(10a\left[L\right]+c)$ functions for various Riemann problems in this study. The Rowdy adaptive activation functions incorporate five trainable parameters, including $a$, $c$, $a_1$, $F_1$, and $c_1$. The $a$ and $c$ trainable coefficients are initialized using a constant value of $0.1$. The $a_1$ coefficient is the amplitude hyper-parameter that is initialized as zero. The phase shift parameter $c_1$ is also initialized with the zero value. The $F_1$ is the frequency hyper-parameter that is initialized with a constant value of $0.1$.

\subsubsection{Positivity preservation for density and pressure}


Ensuring the positivity of density and pressure helps maintain the physical integrity of these conservation laws. Neglecting these constraints can lead to physically unrealistic results and numerical instabilities in simulations and calculations. In this work, we enforce the positivity conditions of pressure and density during the training of the neural network. For each epoch, the predicted density and pressure are forced to be above a small positive value, $\epsilon=10^{-10}$. The same positivity conditions have been enforced in the work of Jagtap et al. \cite{jagtap2022physics}, where they employed physics-informed neural networks to solve inverse problems in supersonic flows; see also \cite{mao2020physics,jagtap2020conservative}, and in the work of Peyvan et al. \cite{PEYVAN2023112310} for hypersonic flow simulations.

\subsection{U-Net with parameter conditioning}

The U-Net is a multiscale convolutional network architecture that is extensively used for solving image segmentation \cite{ronneberger2015u} problems. 
Gupta \emph{et al.} ~\cite{gupta2022towards} introduced the temporal conditioning mechanism that enables the U-Net to learn time-dependent systems. 
The efficiency of this approach was further extended and demonstrated in ~\cite{oommen2023rethinking, ovadia2023ditto}. 
In this work, we extend the same idea and condition the U-Net with respect to the pressure states on the left side of the initial condition profile.

The model consists of two networks. 
1) a U-shaped fully convolutional neural network; and 2) a Multi-Layered Perceptron (MLP). 
The MLP takes $p_l$ as the input and learns a collection of functions $\vec{f}(p_l)$ as the output. The MLP used in this work consists of 2 hidden layers with 128 neurons and learns non-linear functions of $P_l$ as follows:
\begin{equation}
    \vec{f}(p_l) = w^{MLP}_{2} \sin(w^{MLP}_{1} \sin(w^{MLP}_{0} p_l+b^{MLP}_0)+b^{MLP}_1)+b^{MLP}_2,
\end{equation}
where $\{w^{MLP}_i, b^{MLP}_i\}_{i=0}^{2}$ represents the parameters learned during the training of the operator.

The U-Net takes a subset of reference fields from the training dataset ($U^{ref}$) as the input. 
$U^{ref}$ is a collection of discrete representations of density ($\rho$), velocity ($v$) and pressure ($p$) fields sub-sampled from the already available training dataset. 
In this context, $U^{ref}$ is a two-dimensional tensor with $C$ channels and a spatial dimension of $W$. 
The U-Net learns to project $U^{ref}$ to multiple basis functions of the same spatial resolution using the convolutional block that comprises 1) a 1D convolutional layer \cite{krizhevsky2012imagenet}, 2) a group normalization layer \cite{wu2018group}, and 3) a non-linear activation layer. 

The one-dimensional convolution can be performed on a two-dimensional tensor $u$, using a three-dimensional weight tensor $w^{conv}$ with an input channel size of $C$, output channel size of $C'$, and a dimensionality of $W'$, in the following manner:
\begin{equation}
    \text{conv}(u)_{k',i} = \sum_{k=0}^{C-1} \sum_{m=0}^{W'-1}  u_{k, i+m} . w^{conv}_{k,k',m}
\end{equation}
where $w^{conv}$ is the weight tensor learned during the training process.

For the group normalization operation on a two-dimensional tensor $u$, we separate $C$ channels to $G$ groups of $\tilde{C}$ channels ($C=G\times\tilde{C}$), compute $G$ means and standard deviations separately as,
\begin{equation}
    \mu_g = \frac{1}{\tilde{C}.W}\sum_{\tilde{k}=0}^{\tilde{C}-1} \sum_{m=0}^{W-1}  u_{g,\tilde{k},m} \quad \text{, $g$=0,1,...,$G$-1},
\end{equation}
\begin{equation}
    \sigma_g^2 = \frac{1}{\tilde{C}.W} \sum_{\tilde{k}=0}^{\tilde{C}-1} \sum_{m=0}^{W-1}  (u_{g,\tilde{k},m} - \mu_g)^2 \quad \text{, $g$=0,1,...,$G$-1}.
\end{equation}
We then normalize $u$ as,
\begin{equation}
    \hat{u}_{g,\tilde{k},i} = \frac{u_{g,\tilde{k},i} - \mu_g}{\sqrt{\sigma_g^2 + \epsilon}} \quad \text{, $g$=0,1,...,$G$-1} 
\end{equation}
where $\epsilon$ is a small positive constant. 
The output of the group normalization operation is,
\begin{equation}
    \text{GN}(u)_{k,i} = \gamma_{k} . \hat{u}_{k,i} + \beta_k \quad \text{, $k$=0,1,...,$C$-1}.
\end{equation}
Here, $\gamma_k$ and $\beta_k$ are trainable $C$ dimensional parameters to learn the ideal shift and scale operation.

The conv($u$) and GN($u$) are linear transformations of $u$.
We introduce non-linearity using Gaussian Error Linear Unit (GELU) activation function \cite{hendrycks2016gaussian}, which can be approximately represented as, 
\begin{equation}
    \text{GELU}(u) = 0.5 u (1 + \tanh(\sqrt{\frac{2}{\pi}}  (u + 0.044715 u^3))).
\end{equation}
The convolutional block that non-linearly transforms $u$ can be represented as,
\begin{equation}
    \text{conv\_block}(u) = \text{GELU}(\text{GN}(\text{conv}(u))).
\end{equation}

The downsampling operation is performed by the one-dimensional max-pooling layer \cite{yamaguchi1990neural}. 
This operation is expressed as,
\begin{equation}
    \text{down}(u)_{k,i} = \max_{0\leq m < W'}  u_{k,i.S_w+m},
\end{equation}
where $S_w$ represents strides along width and height.
In order to obtain a scale-down factor of 2 during the downsampling operation, $S_w=W'=2$ during maxpooling.
The up-sampling is performed by the one-dimensional transpose convolutional operation \cite{dumoulin2016guide} in the following manner:
\begin{equation}
    \text{up}(u)_{k,i} = \sum_{m=0}^{W'-1} u_{k,i.S_w+m}.w^{tconv}_{m},
\end{equation}
where $w^{tconv}_{m,n}$ is a trainable two-dimensional tensor with dimnesionality $W'$.
To achieve a scale-up factor of 2, $S_w=W'=2$.

Borrowing the notations from \cite{oommen2023rethinking}, the latent representations of $U^{ref}$ can be expressed as, 
\begin{equation}
    \vec{z}^{\mathcal{L}_p} =
    \begin{cases}
         \text{conv\_block}(U^{ref}) & \text{if $p$=1}\\
         \text{conv\_block}(\text{down}(\vec{z}^{\mathcal{L}_{p-1}})) & \text{if $p$=2,3,4}
    \end{cases}
    \label{eq:unet_basis}
\end{equation}
The $\vec{z}^{\mathcal{L}_p}$ can be interpreted as the learned basis functions at 4 different scales. 
We further visualize the eigen spectrum and analyze the eigen modes of the multiscale basis functions in section \ref{subsec:basis_unet}.

Next, we condition the latent basis functions $\vec{z}^{\mathcal{L}_p}$ with the pressure initialization $P_l$ using an element-wise product operation as shown below. 
\begin{equation}
    \vec{z}^{\mathcal{L}_p}_{p_l} = \vec{z}^{\mathcal{L}_p} \odot w^{\mathcal{L}_p}\vec{f}(p_l) \\ \implies (z^{\mathcal{L}_p}_{p_l})_{k,i} = (z^{\mathcal{L}_p})_{k,i} . (w^{\mathcal{L}_p}\vec{f}(p_l))_k \quad \forall p,
\end{equation}
where $w^{\mathcal{L}_p}$ linearly projects $\vec{f}(p_l)$ to number of channels at the $p^{th}$ latent level before conditioning the latent representation. 
We note that the conditioned latent variable $\vec{z}^{\mathcal{L}_p}_{p_l}$ is discrete with respect to space and continuous with respect to the parameter $P_l$.
Next, we upsample the conditioned latent vectors in the following manner,
\begin{equation}
    \vec{d}^{\mathcal{L}_p}_{p_l} =
    \begin{cases}
        \text{up}(\text{conv\_block}(\vec{z}^{\mathcal{L}_{p+1}}_{p_l})) & \text{if $p$=3}\\
        \text{up}(\text{conv\_block}(\vec{z}^{\mathcal{L}_{p+1}}_{p_l}\ocircleplus \vec{d}^{\mathcal{L}_{p}}_{p_l} )) & \text{if $p$=1,2},
    \end{cases}
\end{equation}
where $\ocircleplus$ concatenates the tensors along the dimension of channels.

The output of the model is computed as, 
\begin{equation}
    U(P_l) = \text{GN}(\text{conv}(\text{conv\_block}(\vec{z}^{\mathcal{L}_1}_{p_l}\ocircleplus \vec{d}^{\mathcal{L}_{1}}_{p_l}) )).
\end{equation}
A schematic of the architecture used in this study is presented in Fig.~\ref{fig:unet_arch}. 
 The parameter-conditioned U-Net is trained according to the standard data-driven learning framework.

The parameter conditioned U-Net requires longer training time because of a larger number of parameters (1.5M), consuming 6.3MB of memory. 
Nevertheless, a trained U-Net is fast at inference, and can accurately estimate 100 pressure and velocity fields in 0.17 seconds.
The number of trainable parameters ($n(\theta)$) in a parameter conditioned U-Net is:
\begin{equation}
    n(\theta) = n(\theta_{MLP}) + n(\theta_{norm}) + n(\theta_{conv}),
\end{equation}
where $n(\theta_{MLP})$, $n(\theta_{norm})$ and $n(\theta_{conv})$ represent the number of trainable parameters comprising the multi-layer perception, normalization layers and convolutional layers, respectively. Furthermore, we can approximately represent $n(\theta_{conv})$ as:
\begin{equation}
    n(\theta_{conv}) \propto \sum_{p=1}^{4} K^D C^{(p)} C'^{(p)},
\end{equation}
where $K(=3)$ is the size of the convolution kernel, $D$ is the dimensionality of the problem at hand, $C^{(p)}$ and $C'^{(p)}$ are the number of input and output channels at the $p^{th}$ level of the U-Net. 
Therefore, the U-Net with parameter conditioning can be easily scaled to 2D and 3D problems. 
If we were to use a U-Net architecture with a similar number of channels for a 2D or 3D problem, $n(\theta_{MLP})$ and $n(\theta_{norm})$ would remain unchanged. 
The number of parameters would at most increase by a factor of $K=3$ or $K^2=9$ corresponding to 19MB or 57MB of memory for 2D or 3D problems respectively.     
The efficient, symmetric, and multi-scale architecture with skip connections enables the U-Net to avoid overfitting although it has a large number of training parameters.

\begin{figure}[!h]
  \centering
     \includegraphics[clip, width=\linewidth]{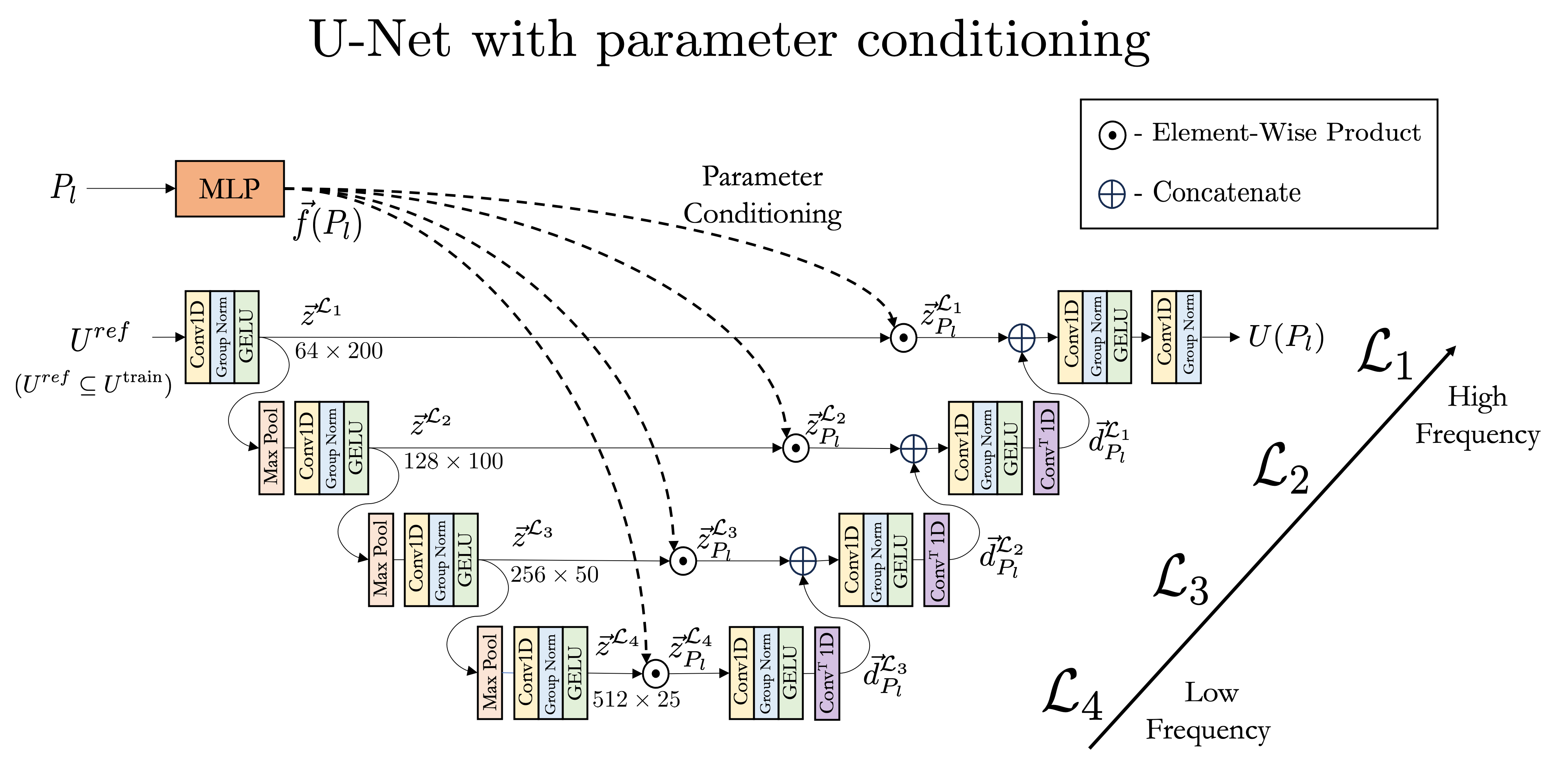}
     \caption{U-Net conditioned on pressure initialization ($p_l$). $U^{ref} \subseteq U^{\text{train}}$ is provided as input to a U-Net which behaves like a multi-scale neural operator. The output of each encoder block, $\vec{z}^{\mathcal{L}_p}$, is conditioned on the parameter $p_l$ through an element-wise product operation. The corresponding representation is concatenated with the previous decoder block's output ($\vec{d}^{\mathcal{L}_{p-1}}_{p_l}$), and subsequently projected back as the output of the model. }
     \label{fig:unet_arch}
\end{figure}


\section{Results}

In the subsequent test cases, we utilize the RiemannONets to establish a mapping from the initial pressure value on the left side of the membrane to the density, velocity, and pressure profiles at a final time. This mapping is performed across 200 equidistant coordinate points situated within the one-dimensional physical domain. The RiemannONets refer to either DeepONet or U-Net architectures. Table~\ref{tbl:cases} presents an overview of the results of this study and provides a comprehensive comparison of several deep neural operators trained and applied for the inference of three Riemann problems. Here, we investigate low-pressure ratio (LPR), intermediate-pressure ratio (IPR), and high-pressure ratio (HPR) Sod problems. For each test case in Table~\ref{tbl:cases}, we train the neural operators with ten different initializations of weights and biases and present the mean and standard deviation of the $L_2$ norm of the error across the ten ensembles. The mean training time of 10 ensembles is also shown in Table~\ref{tbl:cases}. The training time is computed by deploying the code on the NVIDIA RTX 3090 GPUs with Ampere architecture and the on-card memory of 24 GBs. Both branch and trunk nets consist of five hidden layers, each with a width of 150 neurons for the 2-step training DeepONet and 50 for the vanilla DeepONet. Using a single shared trunk neural net, we set up the branch net to infer three variables, including density, velocity, and pressure. The relative $L_2$ norm of the error is computed over the entire test data set. Each sample in the test data set contains 200 density, velocity, and pressure values corresponding to 200 coordinate points. Let $\mathcal{D}$ be a third-order tensor indicating the inferred solution of the test data set with the shape of $\left(N_s, N_p,3\right)$, where $N_s=100$ is the number of samples, $N_p=200$ is the number of points, and 3 referring to the density, velocity, and pressure. The relative $L_2$ norm for each quantity of interest is computed as:

\begin{equation}
L_2\left(E_{k}\right)=\frac{1}{N_s}\sum_{i=1}^{N_s}\frac{\sqrt{\sum_{j=1}^{N_p}\left(D_{i,j,k}-G_{i,j,k}\right)^2}}{\sqrt{(\sum_{j=1}^{N_p}G_{i,j,k}^2)}},\quad k=1,2,3,
    \label{eq:l2norm}
\end{equation}
where $E_k$ denotes the point-wise error of predicted density ($k=1$), velocity ($k=2$), and pressure $k=3$). In Eq.~\eqref{eq:l2norm}, $G_{i,j,k}$ refers to the ground truth for $i^{\textrm{th}}$ test sample at $j^{\textrm{th}}$ coordinate for density, velocity, or pressure solution. The total relative $L_2$ norm of the error is the mean of the $L_2$  norm of error in density, velocity, and pressure.

\begin{table}[!h]
\footnotesize
\caption{Relative L2 norms mean and standard deviation obtained using 10 runs. The $L_2$ norm of the error is calculated over the entire testing dataset for density, velocity, and pressure profiles. The time reported is the training time; the inference time is negligible.}  
\centering
\begin{tabular}{lccccc} 
\hline
Cases &$L_2(\rho)$ \%&$L_2(u)$ \%&$L_2(p)$ \%& total $L_2$ norm \%& Time (Min)\\
\hline
LPR(1 step Tanh, 50) & $0.96\pm0.064$ & $3.92\pm0.251$ & $0.74\pm0.044$& $1.88\pm0.120$ & $12.95$\\
LPR(1 step Tanh, 150) & $1.38\pm0.096$ & $5.45\pm0.371$ & $1.09\pm0.073$& $2.64\pm0.180$ & $12.29$\\
LPR(1 step ReLU, 50) & $1.47\pm0.101$ & $5.74\pm0.408$ & $1.15\pm0.062$& $2.79\pm0.191$ & $11.85$\\
LPR(1 step Rowdy) & $\mathbf{0.70\pm0.103}$ & $\mathbf{2.57\pm0.580}$ & $\mathbf{0.53\pm0.081}$& $\mathbf{1.27\pm0.254}$ & $38.07$\\
\hline\\
LPR(2 step Rowdy) & $\mathbf{0.41\pm0.017}$ & $1.28\pm0.119$ &  $\mathbf{0.33\pm0.047}$ & $0.67\pm0.061$ & $33.45$\\
LPR(U-Net)           & $0.49\pm0.070$ & $\mathbf{0.86\pm0.129}$ &  $0.36\pm0.094$  & $\mathbf{0.57\pm0.098}$ & 1388.58\\
\hline\\
IPR(2 step Rowdy) & $\mathbf{0.33\pm0.027}$ & $\mathbf{0.86\pm0.071}$ & $\mathbf{0.20\pm0.030}$& $\mathbf{0.46\pm0.043}$& 33.06 \\
IPR(U-Net)           & $0.48\pm0.069$ & $0.98\pm0.117$ & $0.50\pm0.156$ & $0.66\pm0.114$ & 1257.57 \\
\hline\\
HPR(2 step Rowdy(QR)) & $0.71\pm0.113$   & $3.50\pm0.248$ & $3.70\pm3.79$& $2.64\pm1.38$& $34.60$ \\
HPR(2 step Rowdy(SVD)) & $\mathbf{0.66\pm0.093}$   & $3.39\pm0.104$ & $2.86\pm1.680$& $2.31\pm0.626$& $26.98$ \\
HPR(U-Net)           & $1.00\pm0.208$ & $\mathbf{2.65\pm0.115}$ & $\mathbf{2.27\pm0.457}$ & $\mathbf{1.97\pm0.260}$ & 1235.67\\
\hline
\end{tabular}
\label{tbl:cases}
\end{table}

In Table~\ref{tbl:cases}, we first compare the accuracy of the vanilla DeepONet framework with adaptive and normal activation functions. We employed the Rowdy adaptive activation function described in Eq.~\eqref{eq:activation} with a $\tanh$ base activation function. For the first comparison, the (LPR) test problem is used. For the vanilla DeepONet, we use weight ($L_2$) regularization and apply density and pressure constraint positivity during training. For all the variables, the vanilla DeepONet with Rowdy activation function performs superior to the DeepONet with $\tanh$ and ReLU activation functions. The training time for the Rowdy activation function is longer than $\tanh$ since the optimizer must minimize the loss function for five additional coefficients of the adaptive activation function besides the weights and biases. In addition,  the forward pass for Rowdy activation functions requires three additional addition operations and evaluation of the $\sin$ terms.

We have also compared the accuracy of one-step Rowdy and two-step Rowdy DeepONet approaches for the LPR case. We can conclude that the two-step training significantly improves the inference accuracy while requiring less training time than the vanilla DeepONet. We can observe that for the LPR case, the U-Net accuracy in inferring velocity and pressure is better than the two-step Rowdy approach. In comparison, the two-step Rowdy provides greater accuracy for density than the U-Net. The training time for the two-step Rowdy approach is significantly less than the U-Net, while the accuracy is comparable. For the IPR case, the two-step Rowdy method obtains better accuracy for all the variables. The two-step Rowdy method provides a small standard deviation for the IPR cases, indicating that the two-step DeepONet is independent of the initialization point. For the HPR test case, we have compared the accuracy of the two-step training approach with SVD and QR-factorization of the trunk net. The SVD factorization demonstrates higher accuracy compared to the QR factorization. Here, we also compare the accuracy of the two-step approach with U-Net. The U-Net provides higher accuracy for velocity and pressure and a higher error for density than the two-step training with the SVD approach. For all the test cases, the computational cost of U-Net is much higher than the two-step training DeepONet approach.

\begin{figure}[!t]
  \begin{center}
    \begin{tabular}{ccc}   \includegraphics[width=0.32\textwidth,height=0.31\textwidth]{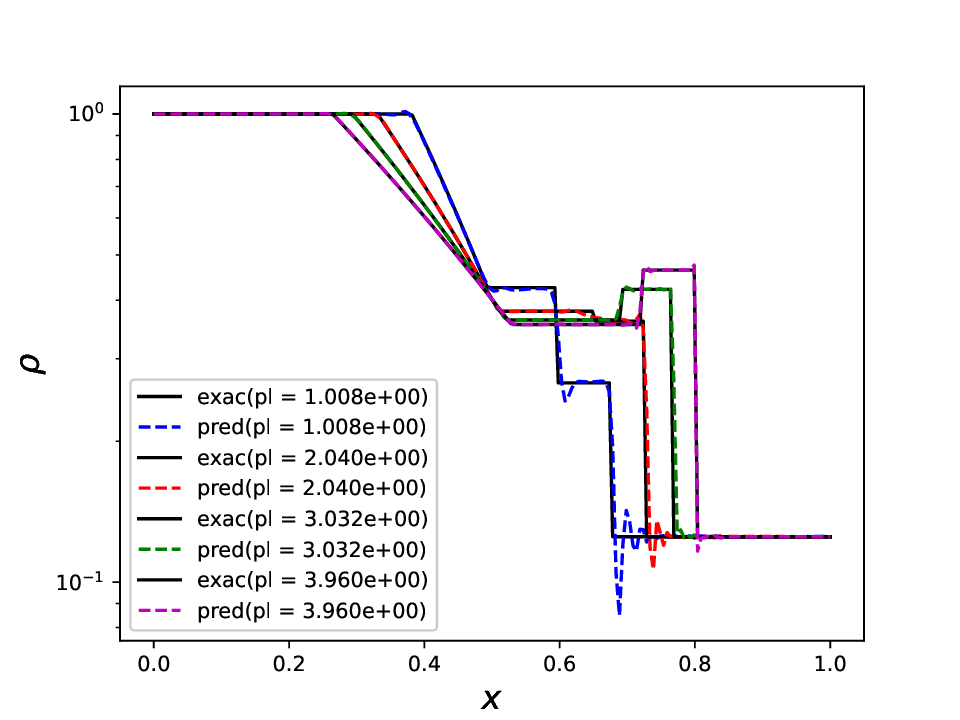}
 &
\includegraphics[width=0.32\textwidth,height=0.31\textwidth]{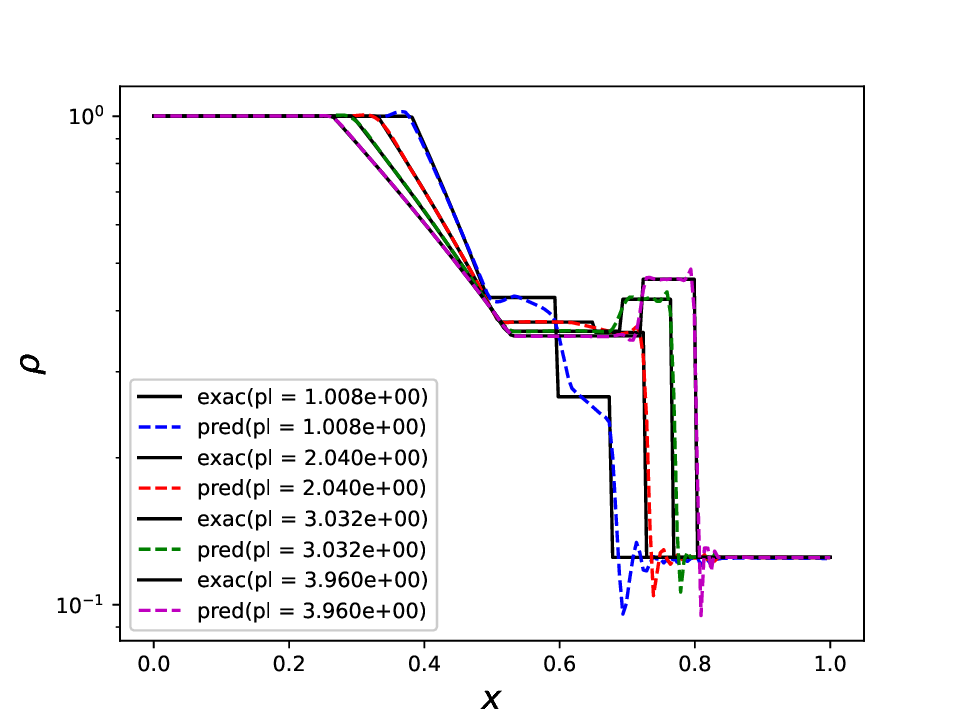} 
 &
\includegraphics[width=0.32\textwidth,height=0.31\textwidth]{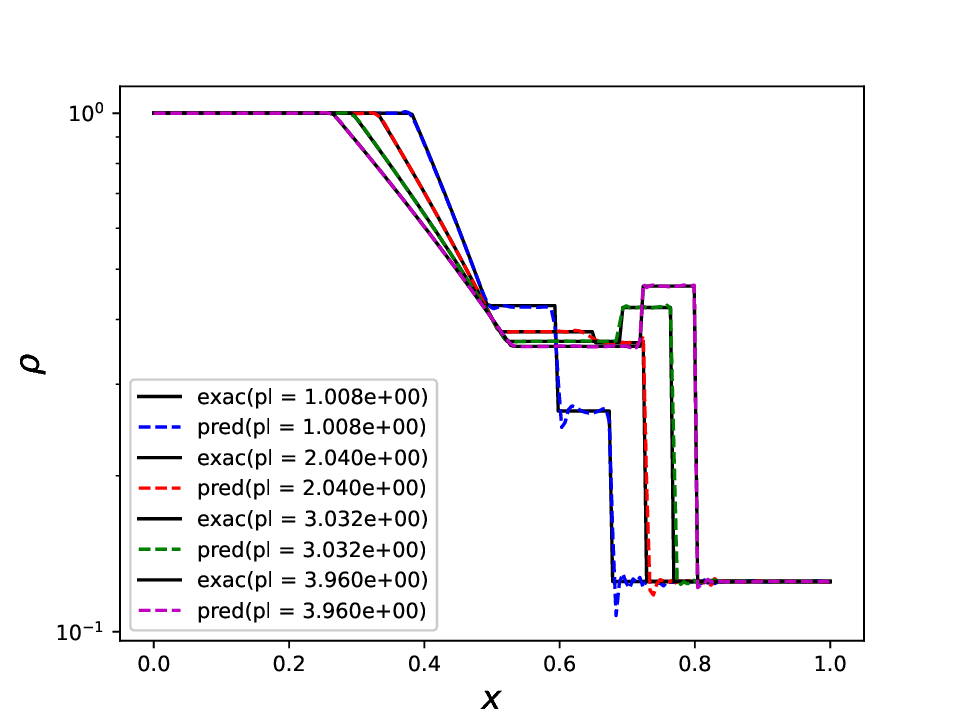}   
  \\
  (a) $\rho_\textrm{test}$, $\tanh$&  (b)$\rho_\textrm{test}$, ReLu& (c) $\rho_\textrm{test}$, Rowdy 
\end{tabular} 
\caption{Low pressure ratio test case: Comparison of $\tanh$, ReLu, and Rowdy $\tanh$ adaptive activation functions. The density of four samples is inferred from the testing data set. The predictive accuracy of the Rowdy  $\tanh$ adaptive activation functions is better than that of its fixed counterparts.}
    \label{fig:case0}
  \end{center}
  
\end{figure}

\subsection{Low pressure ratio Sod problem}
In this case, we solve Eq.~\eqref{eq:general} on a spatial domain defined such as $x \in [0,1]$. The initial conditions for primitive variables are imposed as 

\begin{equation}
\left(\rho, u , p\right)=\begin{cases}
\left(1.0,0.0,p_l\right) & x \le 0.5 \\ 
\left(0.125,0.0,0.1\right) & x > 0.5, 
\end{cases}
\label{eq:low_p}
\end{equation}
where $p_l \in [1.0,5.0]$. We employed the analytical method described in \cite{toro2013riemann} to obtain the results for primitive variables at $t_f=0.1$. From the 500 cases we randomly choose 400 trajectories for training and 100 trajectories for testing data-sets.

 \begin{figure}[!t]
  \begin{center}
    \begin{tabular}{ccc}   \includegraphics[width=0.33\textwidth,height=0.31\textwidth]{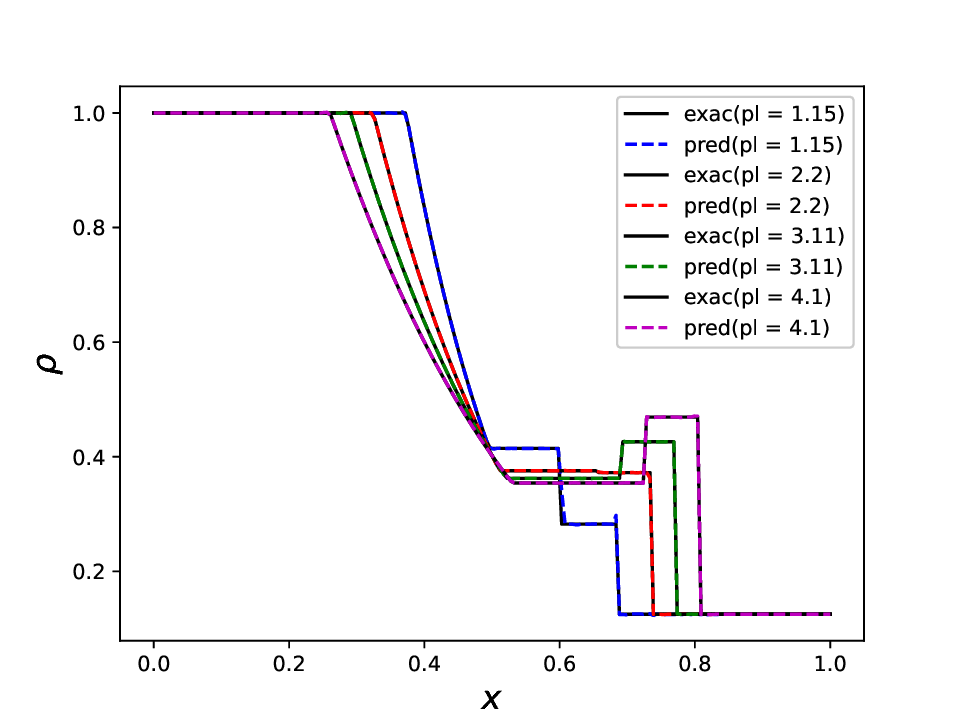}
 &
\includegraphics[width=0.33\textwidth,height=0.31\textwidth]{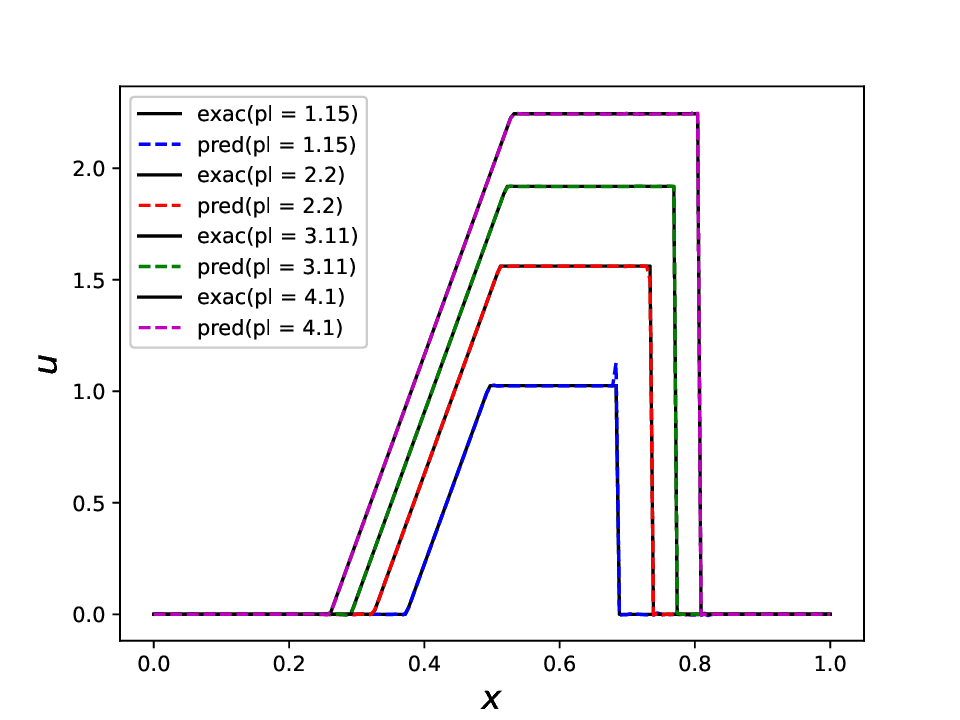}

&
\includegraphics[width=0.33\textwidth,height=0.31\textwidth]{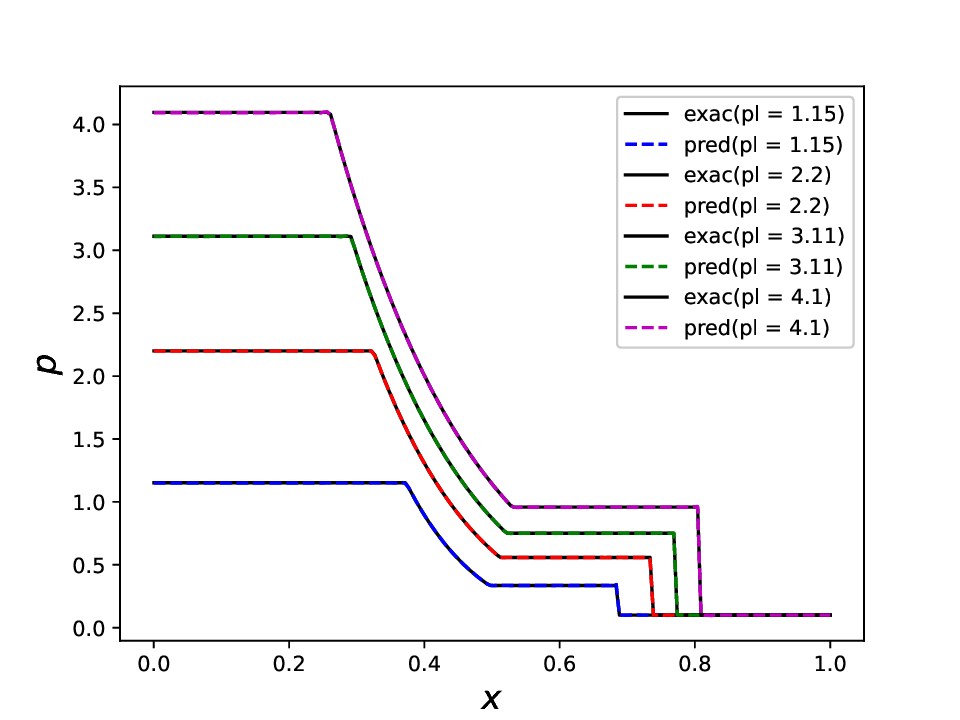}
    
  \\
  (a) $\rho_\textrm{test}$, DeepONet&  (b) $u_\textrm{test}$, DeepONet& (c) $p_\textrm{test}$, DeepONet
\\
\includegraphics[width=0.33\textwidth,height=0.31\textwidth]{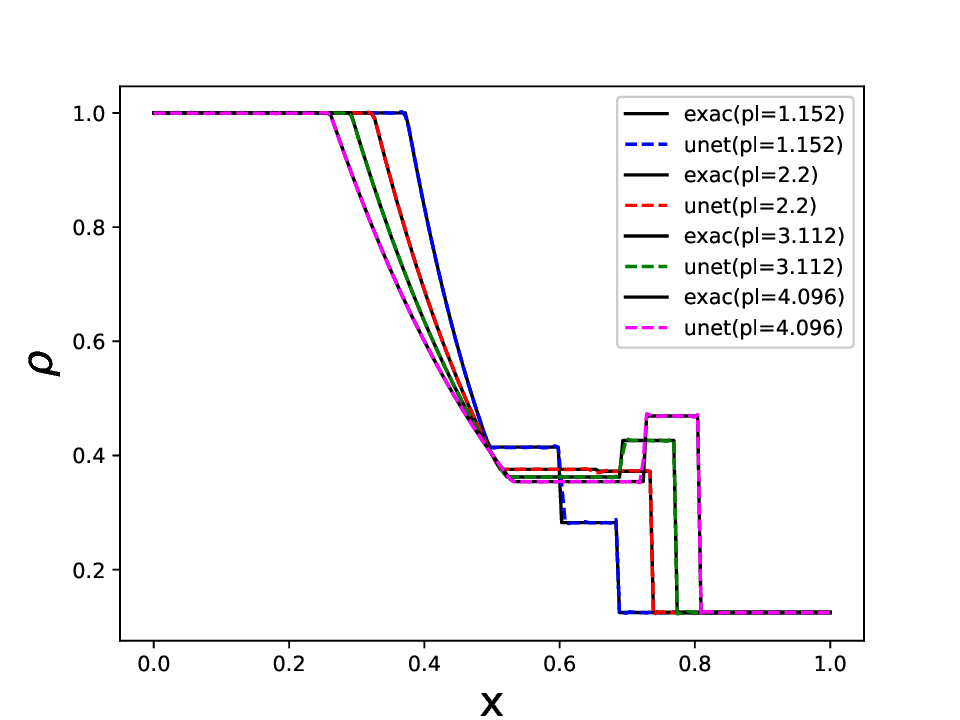}
 &
\includegraphics[width=0.33\textwidth,height=0.31\textwidth]{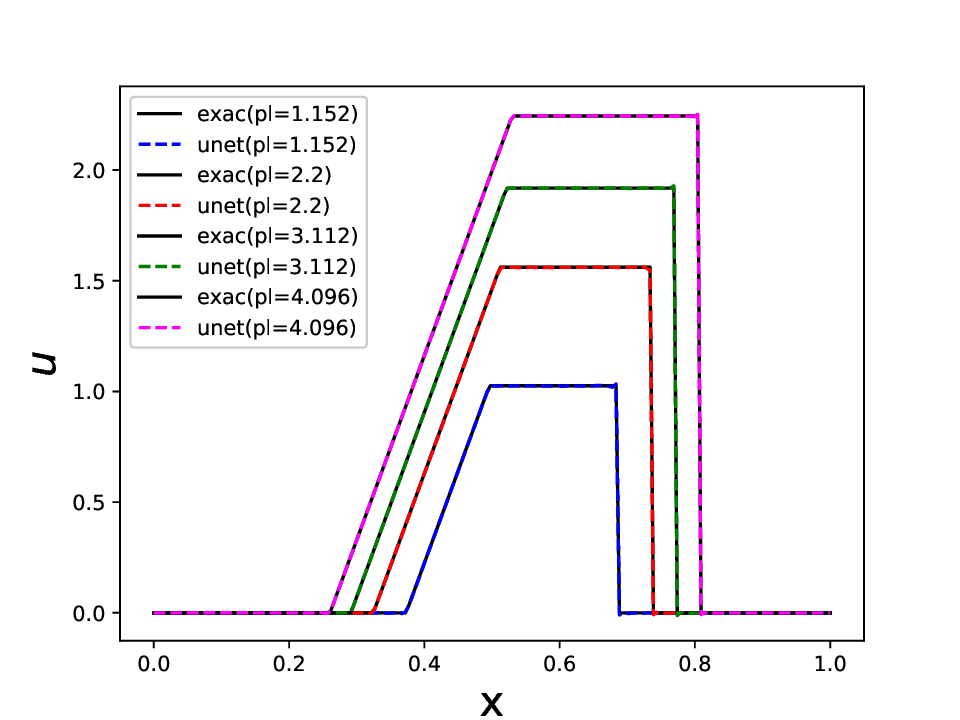}

&
\includegraphics[width=0.33\textwidth,height=0.31\textwidth]{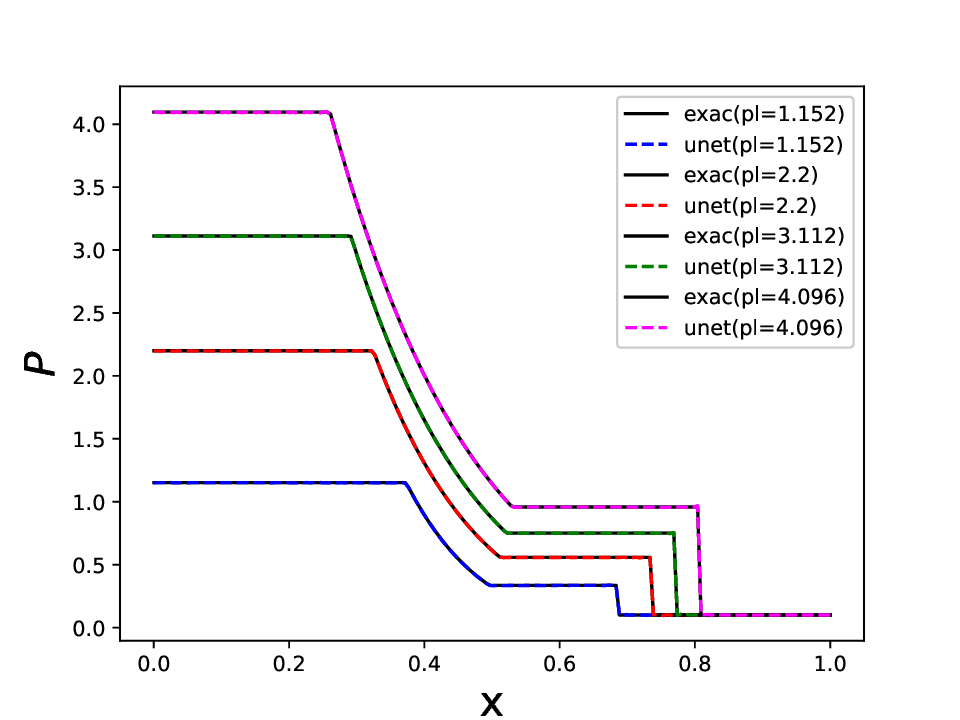}   
  \\
  (d) $\rho_\textrm{test}$, U-Net&  (e) $u_\textrm{test}$, U-Net& (f) $p_\textrm{test}$, U-Net

\end{tabular} 

\caption{Low Pressure Ratio Sod Problem: Comparison of DeepONet and U-Net results. The first row shows the DeepONet results for density, velocity, and pressure, whereas the second row shows the corresponding results for U-Net. The density of four samples is inferred from the testing data set.}
    
    \label{fig:case1}
  \end{center}
  
\end{figure}

We explore the utilization of adaptive activation functions in the DeepONet framework in contrast to conventional $\tanh$ activation functions. Figure \ref{fig:case0} illustrates the testing density results for a low-pressure ratio. Fix activation functions are employed in Fig.~\ref{fig:case0}(a) and (b), while the adaptive Rowdy activation function is used for Fig.~\ref{fig:case0}(c). It is evident that, compared to the Rowdy activation functions, the fixed activation function introduces more oscillations, particularly in the vicinity of discontinuous solutions such as contact and shock waves. Despite the increased computational cost associated with Rowdy activation, relying solely on fixed activation functions for training is unlikely to enhance the accuracy of predictions. Henceforth, we have employed the Rowdy activation function for all subsequent test cases.

The results of the DeepONet and U-Net models are illustrated in Figure \ref{fig:case1}. Specifically, the first row displays the outcomes of DeepONet for density, velocity, and pressure, while the second row presents the corresponding results for U-Net. The density for four distinct samples is inferred from the testing dataset. Notably, both DeepONet and U-Net exhibit good performance, particularly in addressing the challenges posed by the low-pressure ratio problem. In particular, while capturing the shock, contact, and expansion waves, both architectures exhibit negligible overshoots and undershoots. Focusing on the density profiles, we can observe that the U-Net provides some low-amplitude oscillations in the blue curve compared to the DeepONet counterpart. However, DeepONet induces an overshoot in the velocity profile of $pl=1.15$ at the shock wave location. The visual analysis from the figure matches the $L_2$ norm results shown in Table~\ref{tbl:cases}.




\begin{figure}[!t]
  \begin{center}
    \begin{tabular}{ccc}   \includegraphics[width=0.33\textwidth,height=0.31\textwidth]{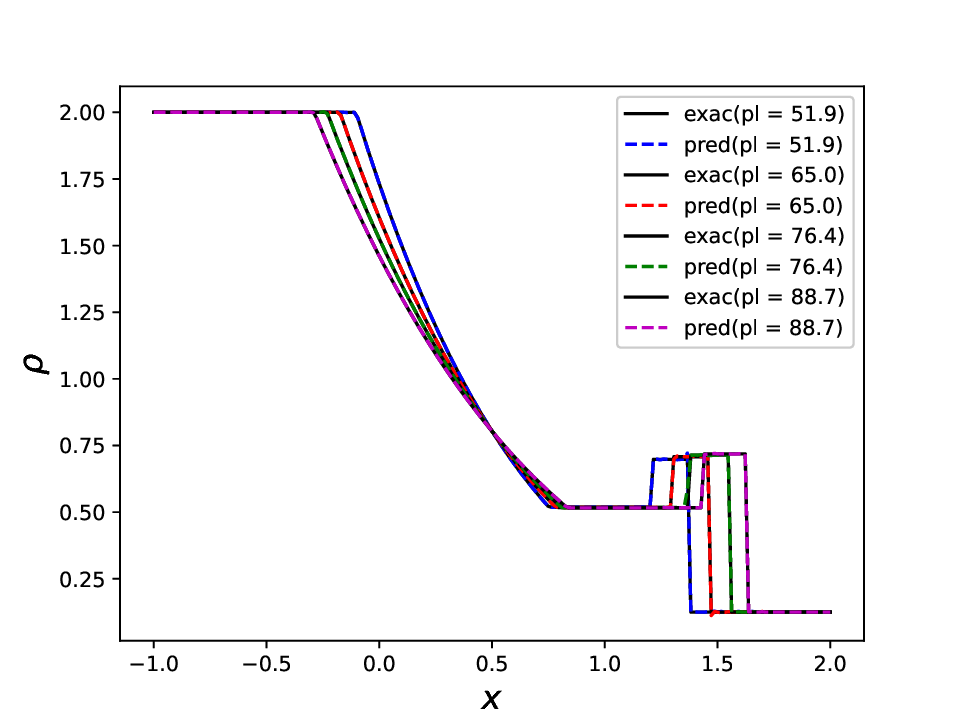}
 &
\includegraphics[width=0.33\textwidth,height=0.31\textwidth]{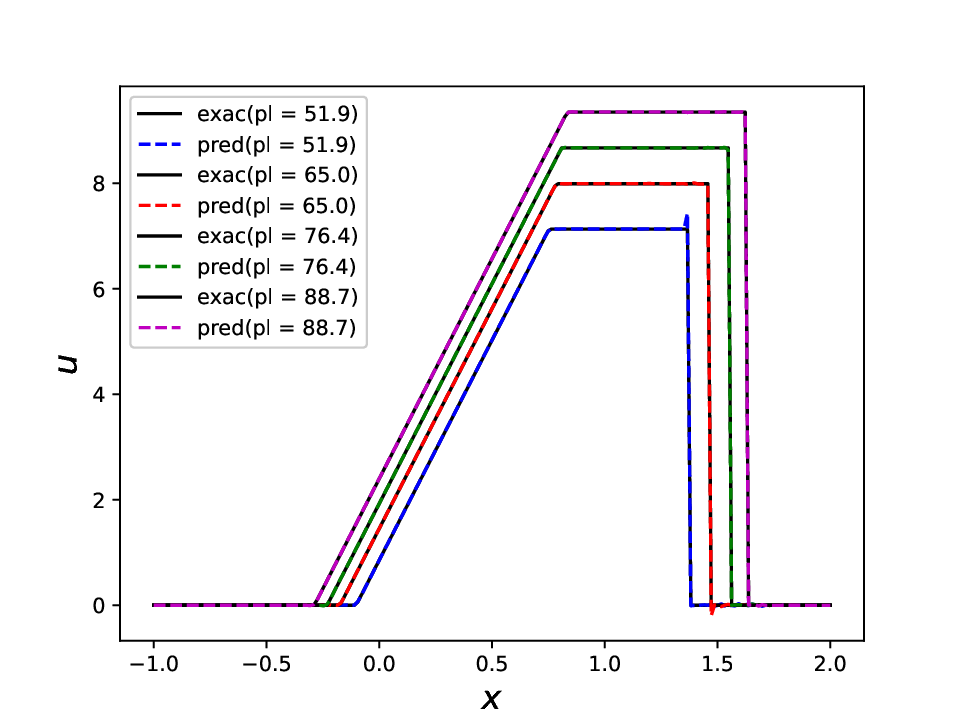}

&
\includegraphics[width=0.33\textwidth,height=0.31\textwidth]{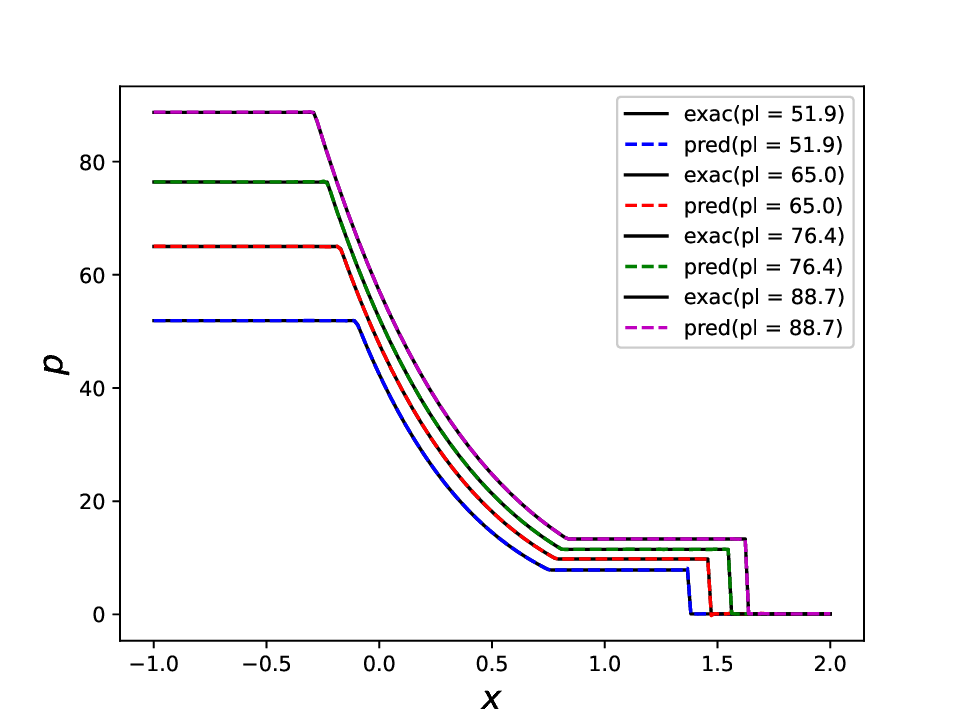}
    
  \\
  (a) $\rho_\textrm{test}$, DeepONet&  (b) $u_\textrm{test}$, DeepONet& (c) $p_\textrm{test}$, DeepONet
\\
\includegraphics[width=0.33\textwidth,height=0.31\textwidth]{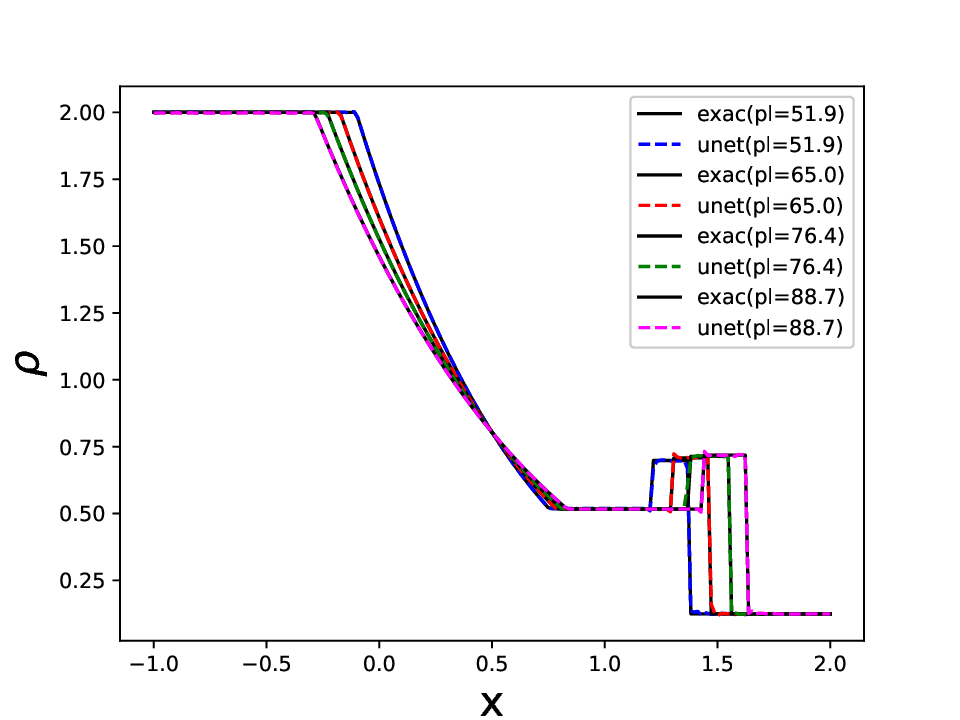}
 &
\includegraphics[width=0.33\textwidth,height=0.31\textwidth]{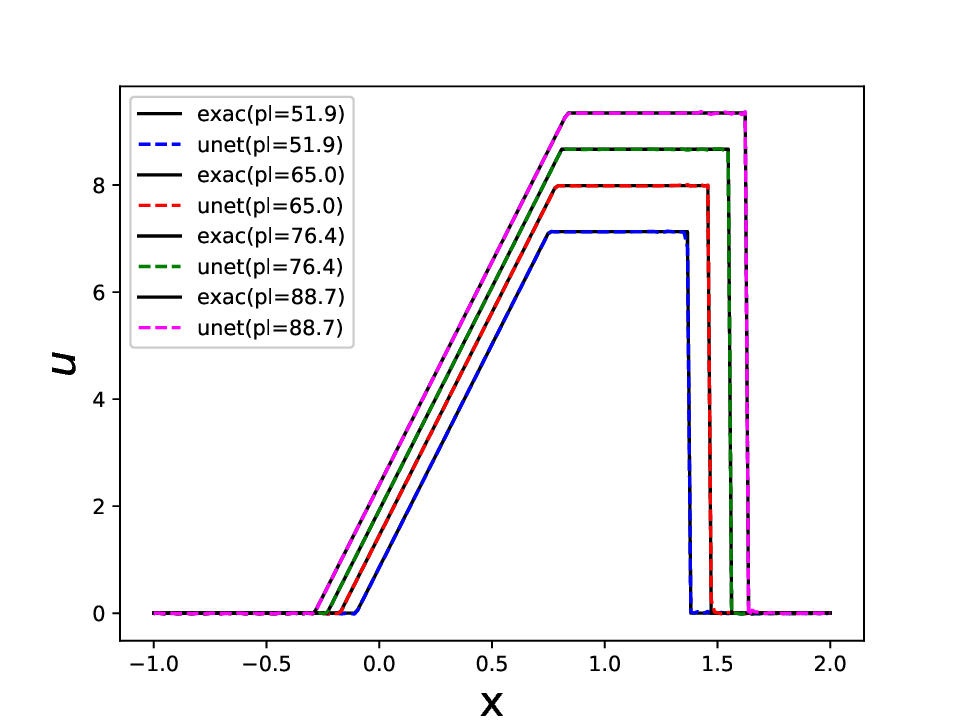}

&
\includegraphics[width=0.33\textwidth,height=0.31\textwidth]{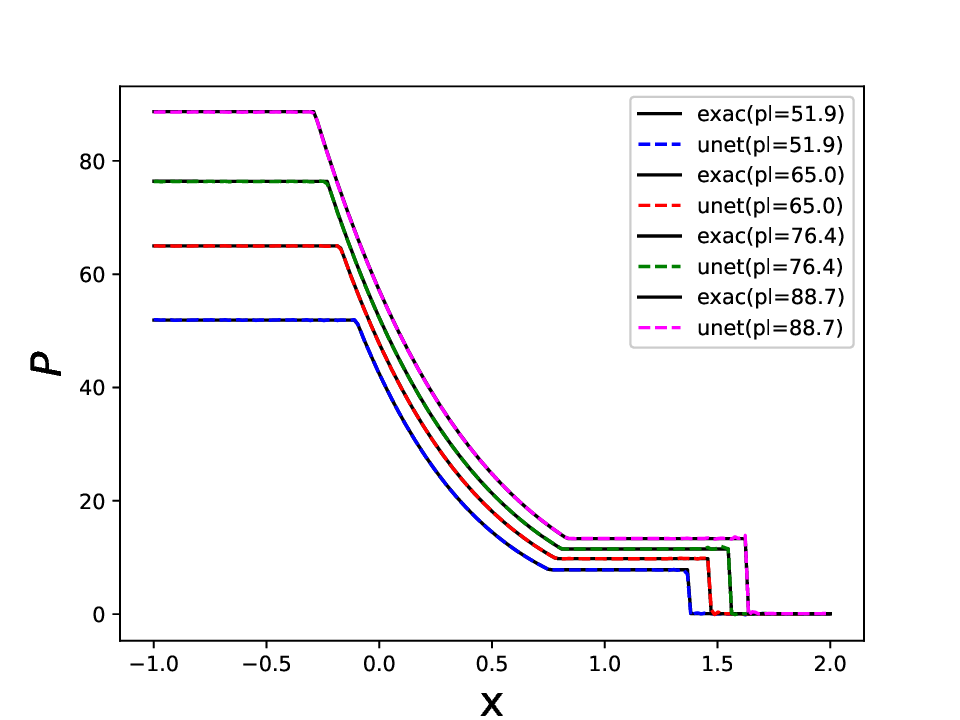}   
  \\
  (d) $\rho_\textrm{test}$, U-Net&  (e) $u_\textrm{test}$, U-Net& (f) $p_\textrm{test}$, U-Net

\end{tabular} 

\caption{Intermediate Pressure Ratio Sod Problem: Comparison of DeepONet and U-Net results. The first row shows the DeepONet results for density, velocity, and pressure, whereas the second row shows the corresponding results for U-Net. The density of four samples is inferred from the testing data set.}
    
    \label{fig:case2}
  \end{center}
  
\end{figure}

\subsection{Medium pressure ratio Sod problem}

Next, we select a more challenging Sod problem for which we increase the pressure and density ratio. We solve Eq.~\eqref{eq:general} on a spatial domain defined as $x \in [-1,2]$. For this case, the initial conditions are selected as 

\begin{equation}
\left(\rho, u , p\right)=\begin{cases}
\left(2.0,0.0,p_l\right) & x \le 0.5 \\ 
\left(0.125,0.0,0.1\right) & x > 0.5, 
\end{cases}
\label{eq:low_p}
\end{equation}
where $p_l \in [50.0,100.0]$. We again use the exact method to obtain the results at $t_f=0.1$. We choose 500 equispaced various $p_l$ values and randomly assign 400 cases for training and 100 for testing.

Figure \ref{fig:case2} depicts the outcomes obtained from DeepONet and U-Net for the IPR problem, showcasing the testing results for density, velocity, and pressure. The initial row presents the results of DeepONet for density, velocity, and pressure, while the subsequent row displays the corresponding outcomes for U-Net. The density values for four distinct samples are extrapolated from the testing data set. Notably, in this instance, the accuracy of DeepONet is slightly better than that of U-Net, particularly evident in the absence of oscillations near the shock wave location at the right-most discontinuities in the curves. 

\begin{figure}[h!]
  \begin{center}
    \begin{tabular}{ccc}   \includegraphics[width=0.33\textwidth,height=0.31\textwidth]{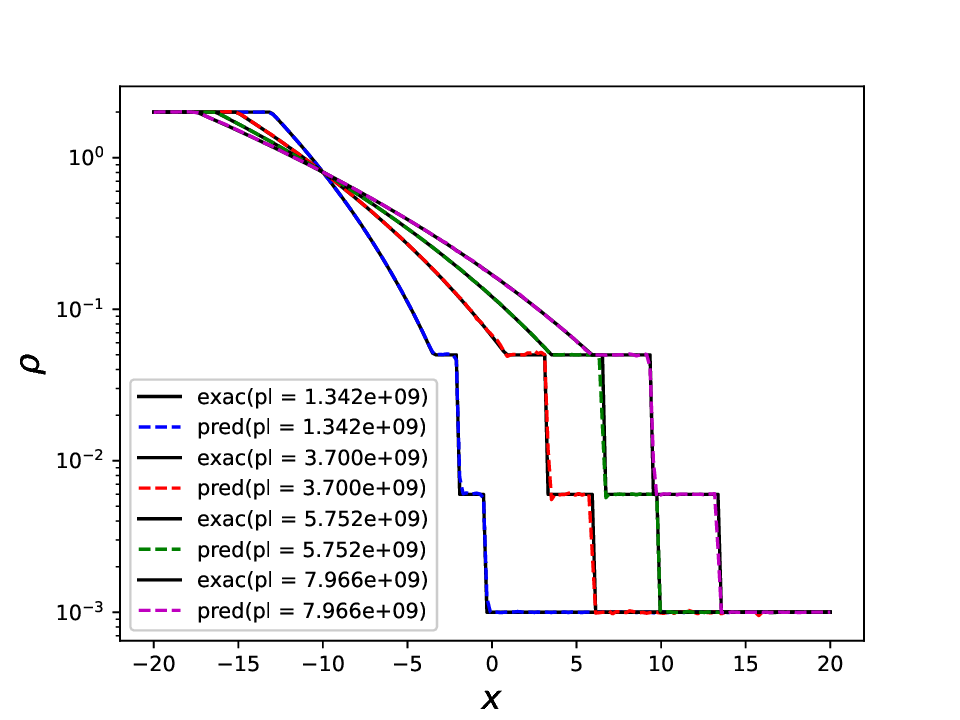}
 &
\includegraphics[width=0.33\textwidth,height=0.31\textwidth]{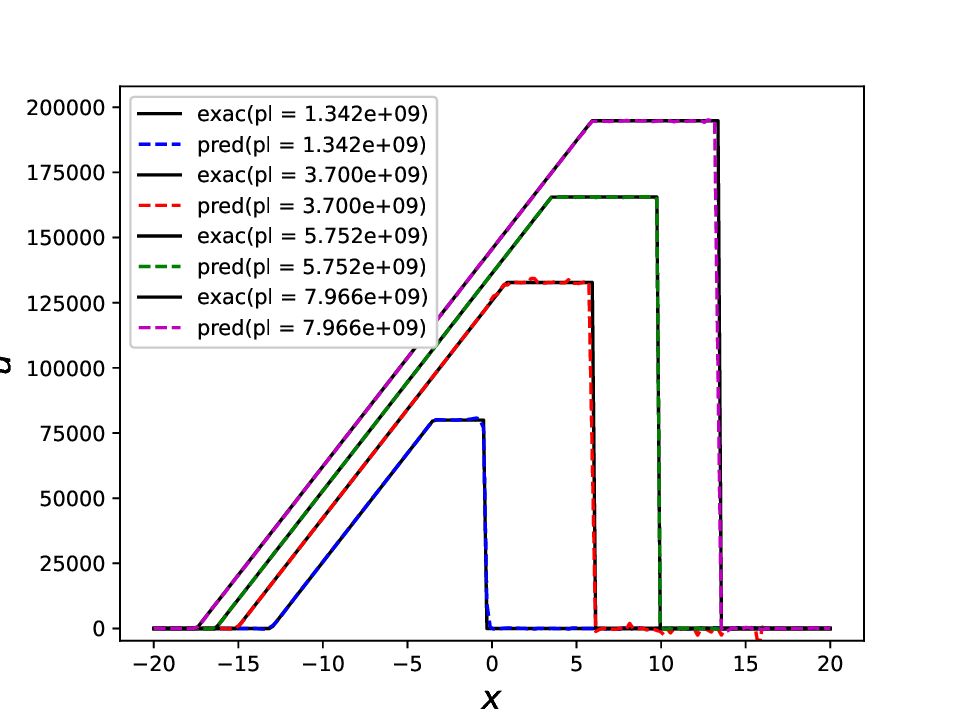}

&
\includegraphics[width=0.33\textwidth,height=0.31\textwidth]{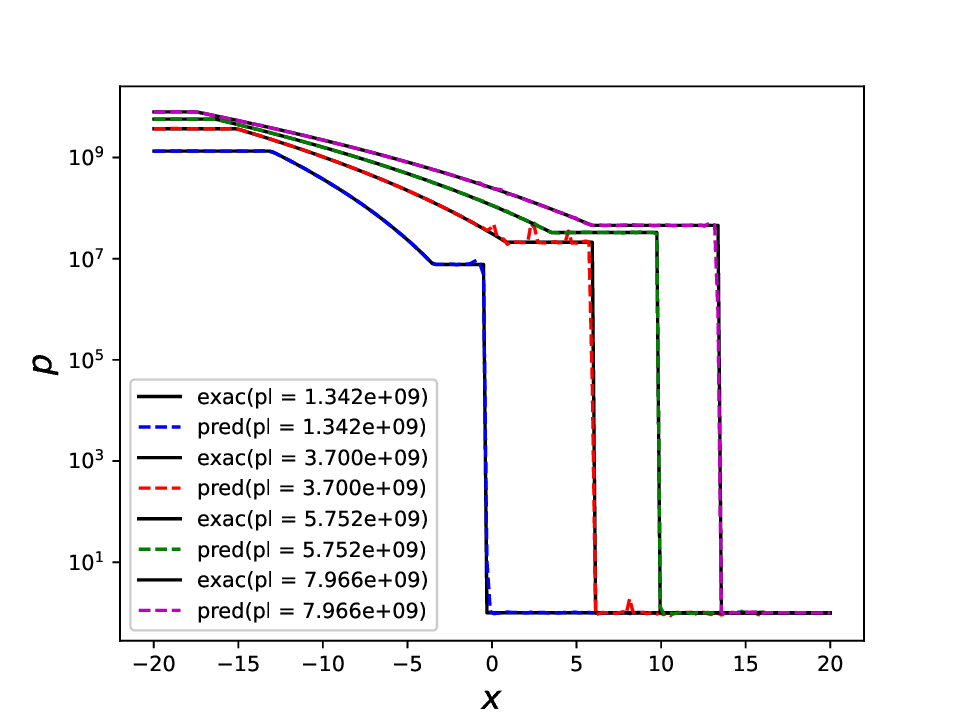}
    
  \\
  (a) $\rho_\textrm{test}$, DeepONet&  (b) $u_\textrm{test}$, DeepONet& (c) $p_\textrm{test}$, DeepONet
\\
\includegraphics[width=0.33\textwidth,height=0.31\textwidth]{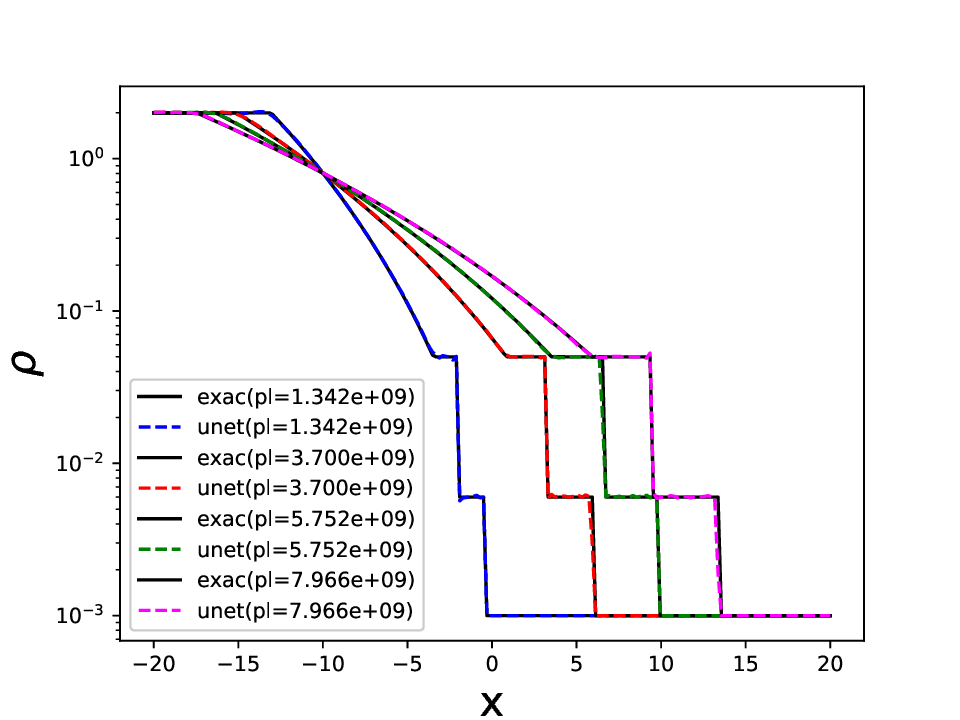}
 &
\includegraphics[width=0.33\textwidth,height=0.31\textwidth]{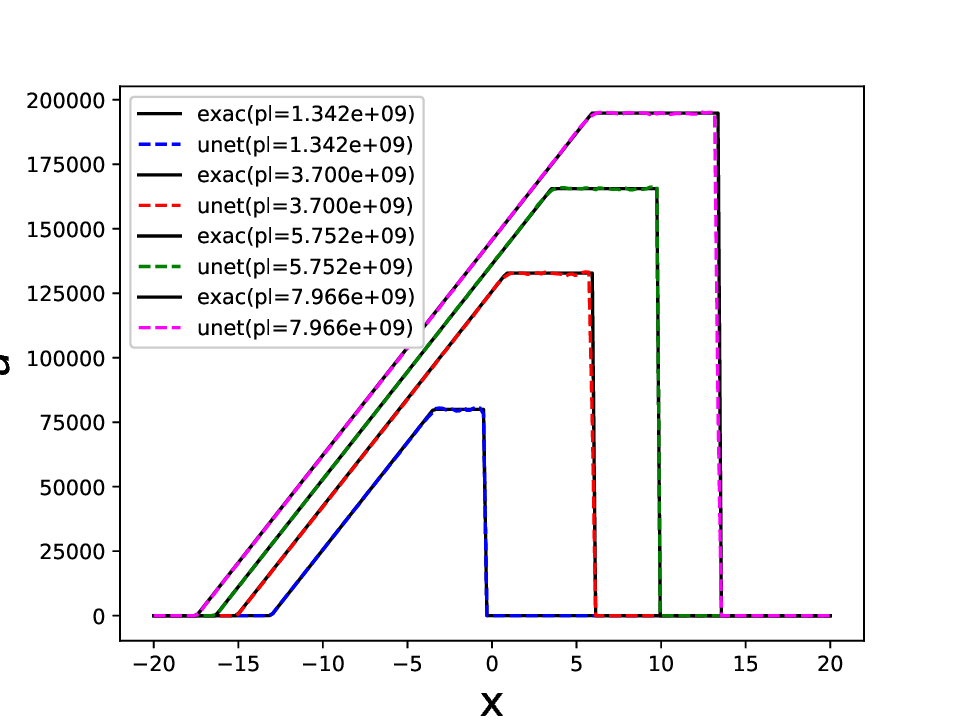}

&
\includegraphics[width=0.33\textwidth,height=0.31\textwidth]{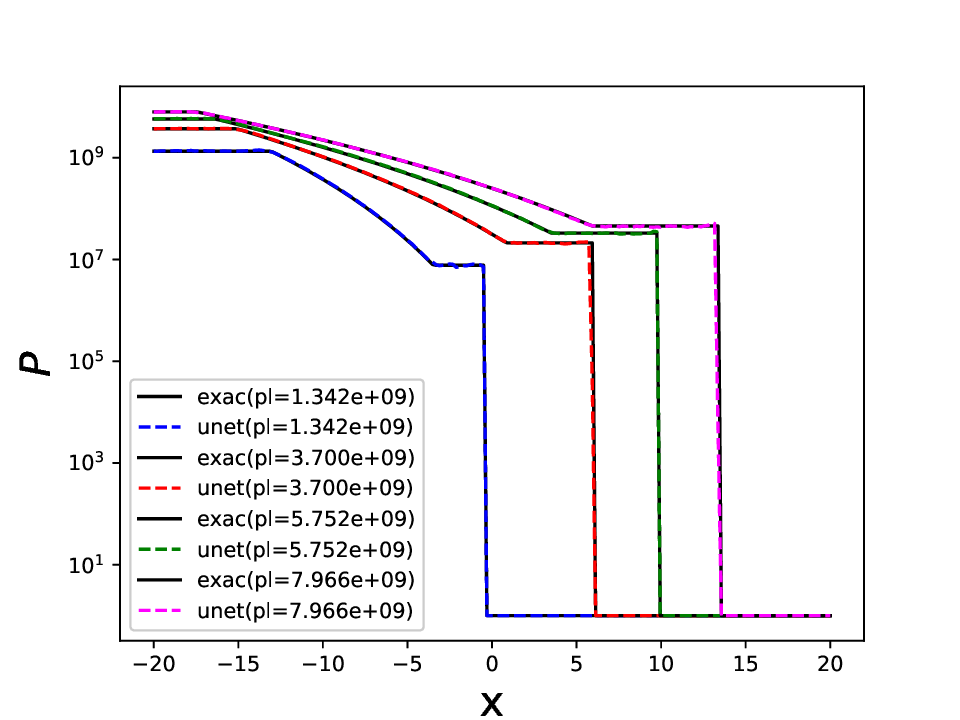}   
  \\
  (d) $\rho_\textrm{test}$, U-Net&  (e) $u_\textrm{test}$, U-Net& (f) $p_\textrm{test}$, U-Net

\end{tabular} 

\caption{High Pressure Ratio Sod Problem: Comparison of DeepONet and U-Net results. The first row shows the DeepONet results for density, velocity, and pressure, whereas the second row shows the corresponding results for U-Net. The density of four samples is inferred from the testing data set.}
    
    \label{fig:case3}
  \end{center}
  
\end{figure}
\subsection{High pressure ratio Sod problem (LeBlanc problem)}
For the last test case, we increase the pressure ratio to the extreme values employed in the so-called LeBlanc-Sod problem. We solve Eq.~\eqref{eq:general} on a spatial domain defined as $x \in [-20,20]$. For this case, the initial conditions are selected as 

\begin{equation}
\left(\rho, u , p\right)=\begin{cases}
\left(2.0,0.0,p_l\right) & x \le -10 \\ 
\left(0.001,0.0,1.0\right) & x > -10, 
\end{cases}
\label{eq:low_p}
\end{equation}
where $p_l \in [10^9,10^{10}]$. We use the exact method to obtain the results at $t_f=0.0001$. We choose 500 equispaced various $p_l$ values and randomly assign 400 cases for training and 100 for testing. For training the neural nets on this problem, we employ the logarithm of density and pressure values to construct the loss function. For the inference stage, we apply the exponential function on the density and pressure to convert the predicted values to physical values of pressure and density.

The LeBlanc problem poses a significant challenge due to a very high-pressure ratio. Figure \ref{fig:case3} presents the inferred results of DeepONet and U-Net for density, velocity, and pressure. Notably, these results highlight the testing outcomes. U-Net adeptly captures discontinuous features such as shocks and contacts with higher accuracy (without oscillations) than the DeepONet.

\section{Operator Representation}
This section is dedicated to the analysis of data-driven basis functions, forming the foundation for obtaining representations of the two neural operators. It's worth highlighting that the basis for DeepOnet is continuous, while for U-Net, it takes a discrete form.

\begin{figure}[!h]
  \begin{center}
    \begin{tabular}{ccc}   \includegraphics[width=0.33\textwidth,height=0.31\textwidth]{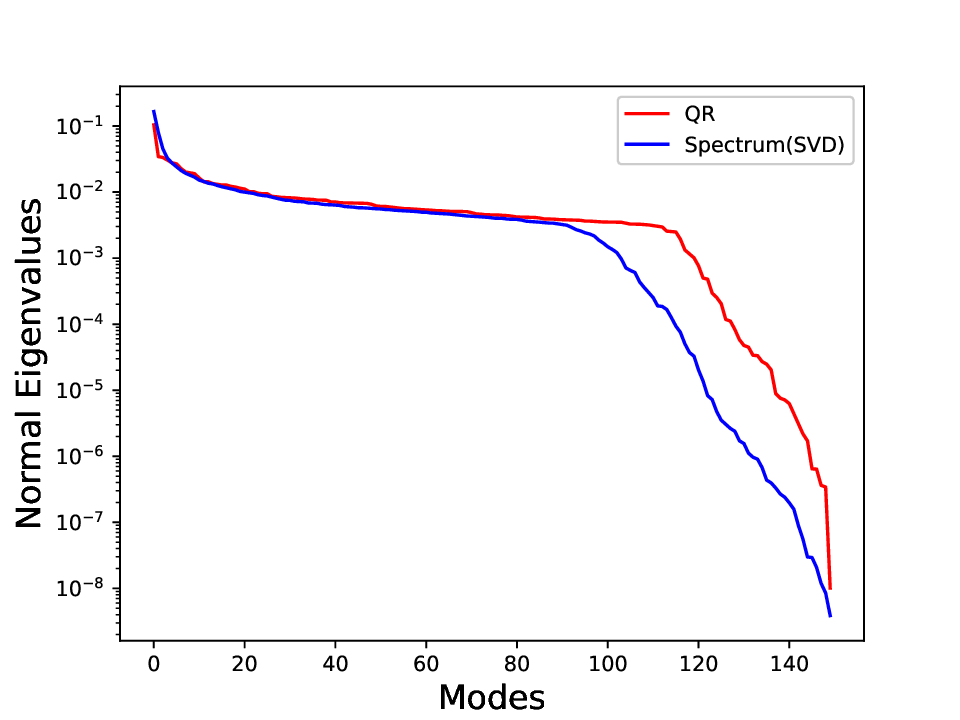}
 &
\includegraphics[width=0.33\textwidth,height=0.31\textwidth]{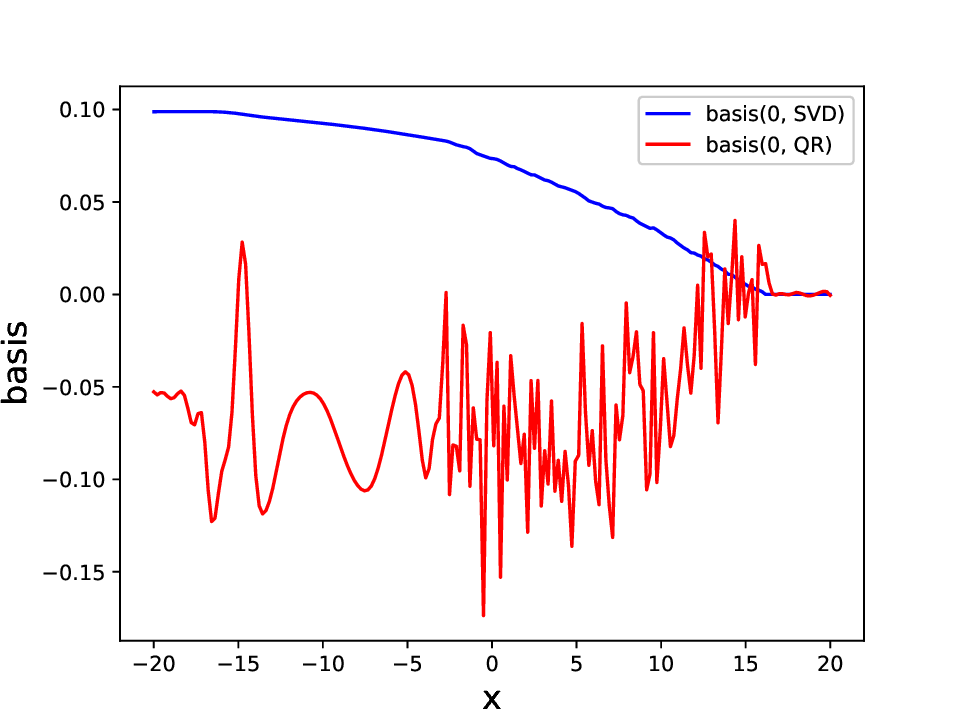}

&
\includegraphics[width=0.33\textwidth,height=0.331\textwidth]{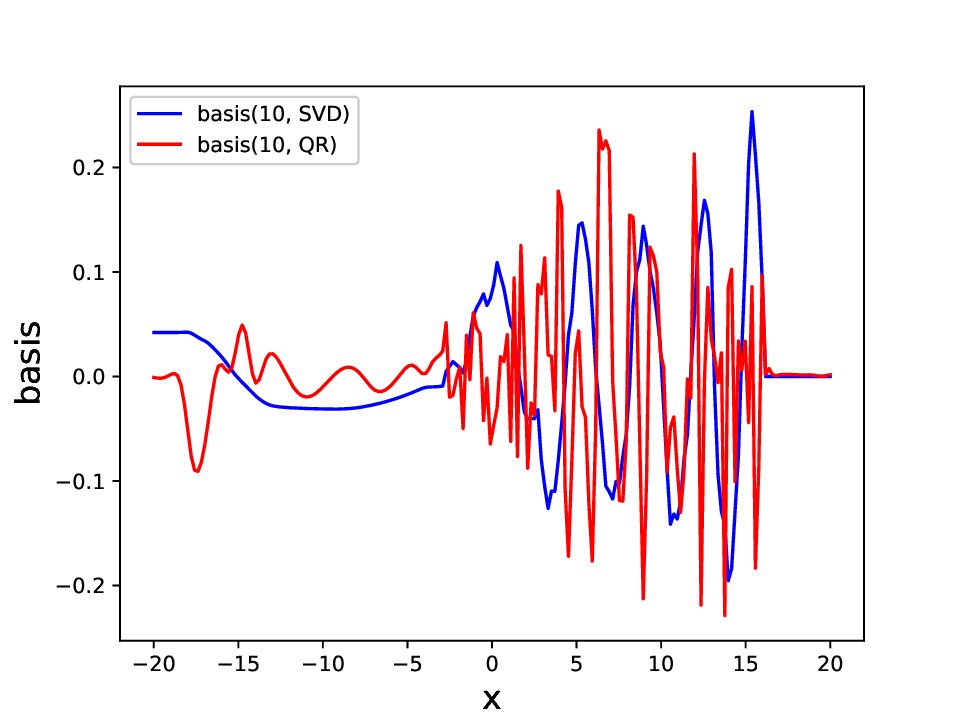}
    
  \\
  (a) Spectrum&  (b) Mode$=0$ basis& (c) Mode$=10$ basis
\\
  \includegraphics[width=0.33\textwidth,height=0.31\textwidth]{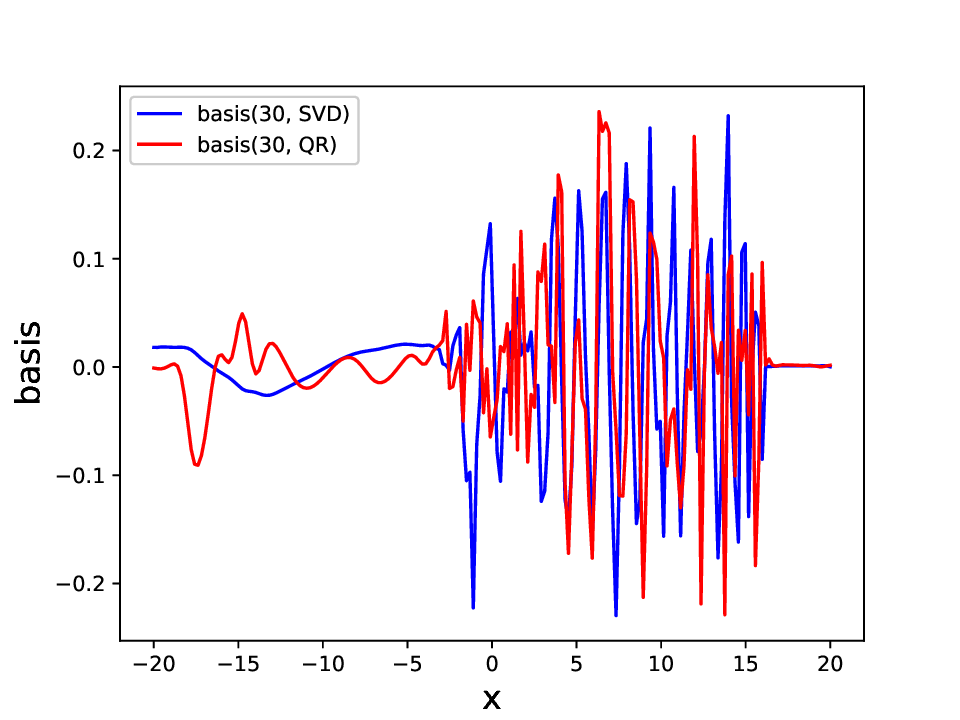}
 &
\includegraphics[width=0.33\textwidth,height=0.31\textwidth]{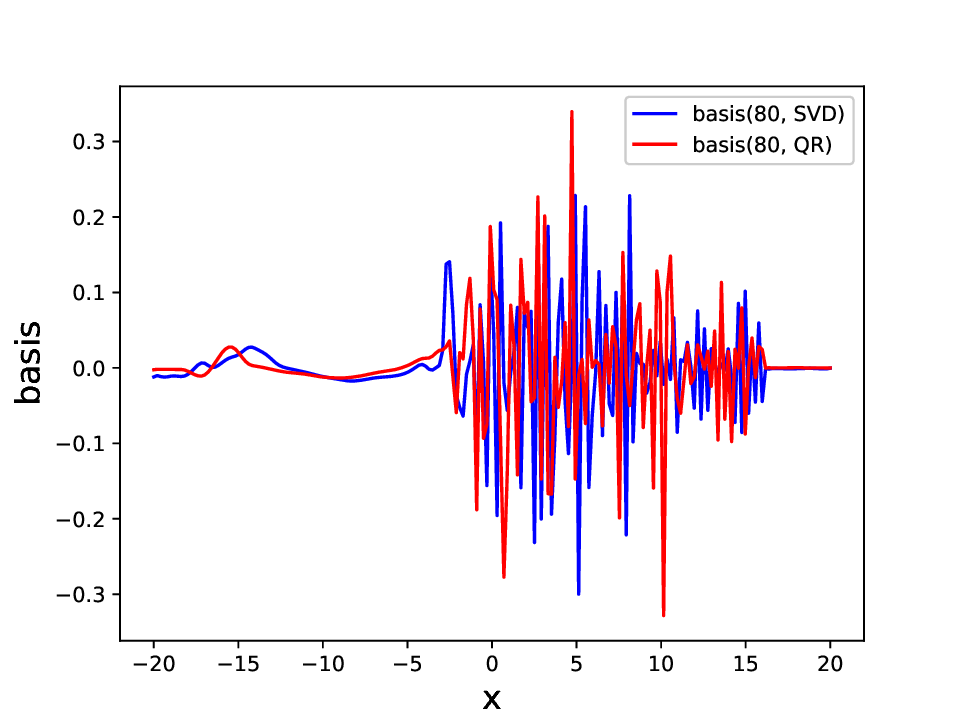}
 &
\includegraphics[width=0.33\textwidth,height=0.31\textwidth]{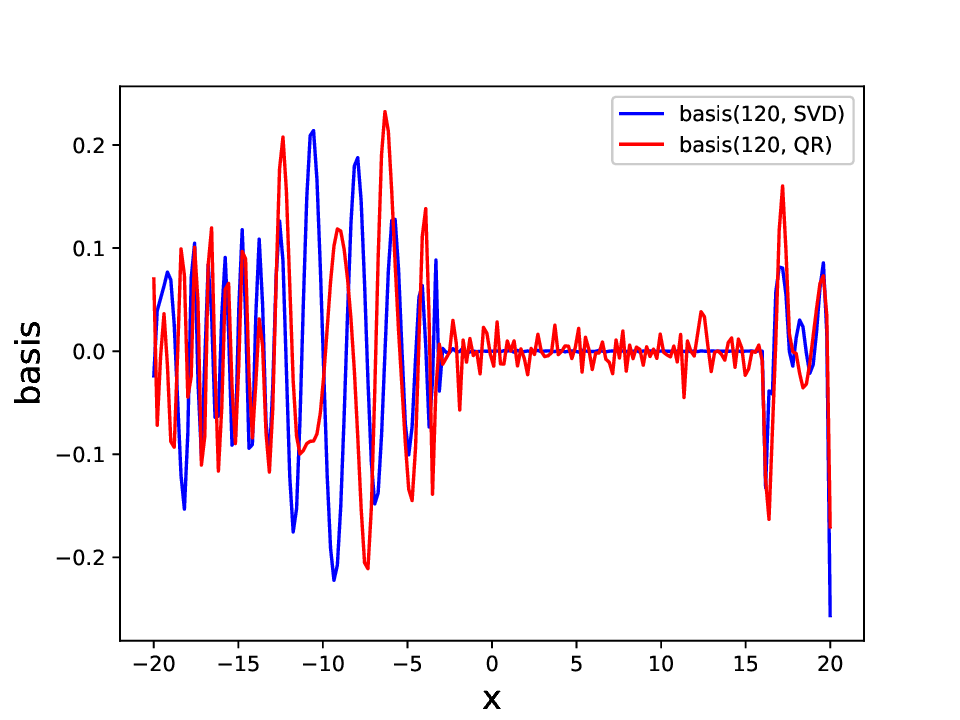}
  \\
  (d) Mode$=30$ basis&  (e) Mode$=80$ basis&  (f) Mode$=120$ basis

\end{tabular} 

\caption{QR vs. SVD spectrum and basis comparison: The top-left plot shows the spectrum comparison between QR and SVD. The rest of the plots  compare the different modes of QR and SVD.}
    
    \label{fig:compsvdvsqr}
  \end{center}
  
\end{figure}

\subsection{Basis functions - DeepONet}

In Fig.~\ref{fig:compsvdvsqr} we compare the spectrum of singular values obtained using QR and SVD decomposition of the trunk net output using the optimized trunk parameters for the training of the HPR problem. In QR decomposition, the contribution of columns of matrix $Q$ in forming the trunk net output matrix is determined by calculating the $L_2$ norm of the rows of matrix $R$. This norm represents the energy captured by each eigenvector, i.e., the columns of matrix $Q$. We sorted the eigenvectors based on their amount of energy. Some of the basis functions for five different modes are depicted for comparison. We can observe that the SVD basis functions exhibit a hierarchical structure from the lowest to highest modes, where we observe high-frequency oscillations in the basis function profile. In contrast, the basis functions computed using the QR factorization are not hierarchical since mode zero exhibits high-frequency features like the higher modes.
Moreover, the QR spectrum exhibits high energy across a high range of modes, inferring that the contribution of all the modes is equal. Both sets of 150 eigenfunctions obtained from QR and SVD decomposition are orthonormal; however, the spectrum of SVD approaches zero at the highest mode faster, indicating a smaller amount of oscillations than the QR spectrum. Therefore, the SVD factorization is more robust than the QR factorization and can be used for feature extraction since the low-mode eigenfunctions mimic the physical shape of the solution profiles. 

\begin{figure}[!h]
  \begin{center}
    \begin{tabular}{ccc}   \includegraphics[width=0.33\textwidth,height=0.31\textwidth]{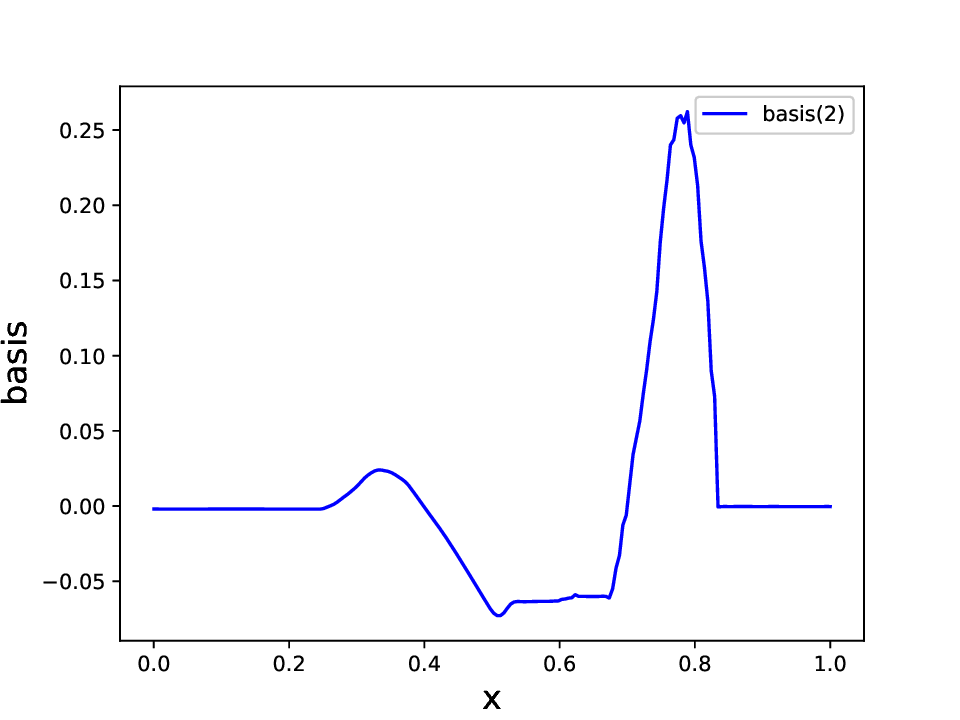}
 &
\includegraphics[width=0.33\textwidth,height=0.31\textwidth]{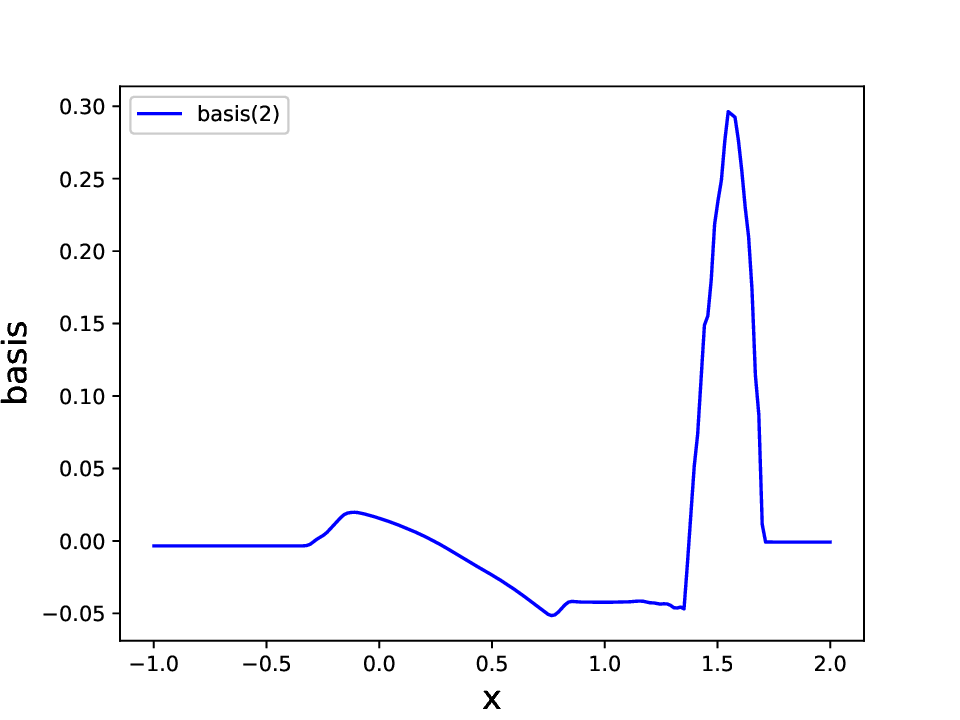}

&
\includegraphics[width=0.33\textwidth,height=0.31\textwidth]{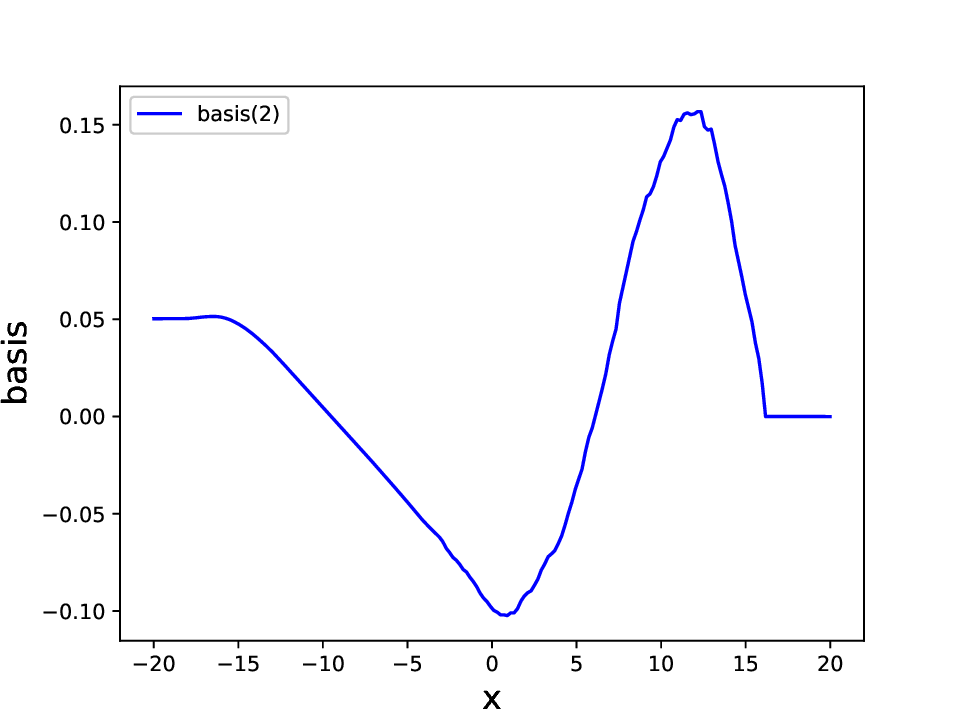}
    
  \\
  (a) LPR&  (b) IPR& (c) HPR

\end{tabular} 

\caption{Effect of jump in pressure ratio: We compare the third basis for low (a), intermediate (b), and high pressure (c) ratios.}
    
    \label{fig:basis_jump}
  \end{center}
  
\end{figure}

In Fig.~\ref{fig:basis_jump}, we compare the third mode basis functions for three problems from low to high-pressure ratios. The pressure ratio range varies such that $p_l\in[10,50]$, $p_l\in[500,1000]$, and $p_l\in[10^9,10^{10}]$ for LPR, IPR, and HPR problems, respectively. We can observe that the shape of the low modes remains approximately the same from LPR to IPR cases, indicating that the data-driven basis function obtained using the low-pressure training data set can train a branch net for the intermediate pressure data set for at least the low modes. 

\begin{figure}[!t]
  \begin{center}
    \begin{tabular}{ccc}   \includegraphics[width=0.33\textwidth,height=0.31\textwidth]{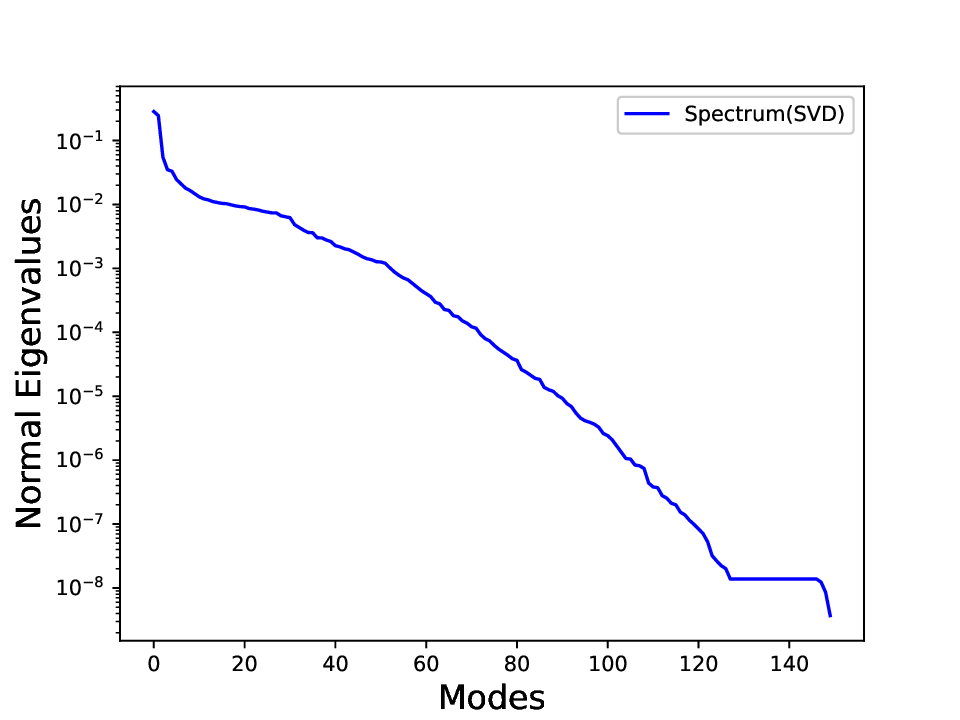}
 &
\includegraphics[width=0.33\textwidth,height=0.31\textwidth]{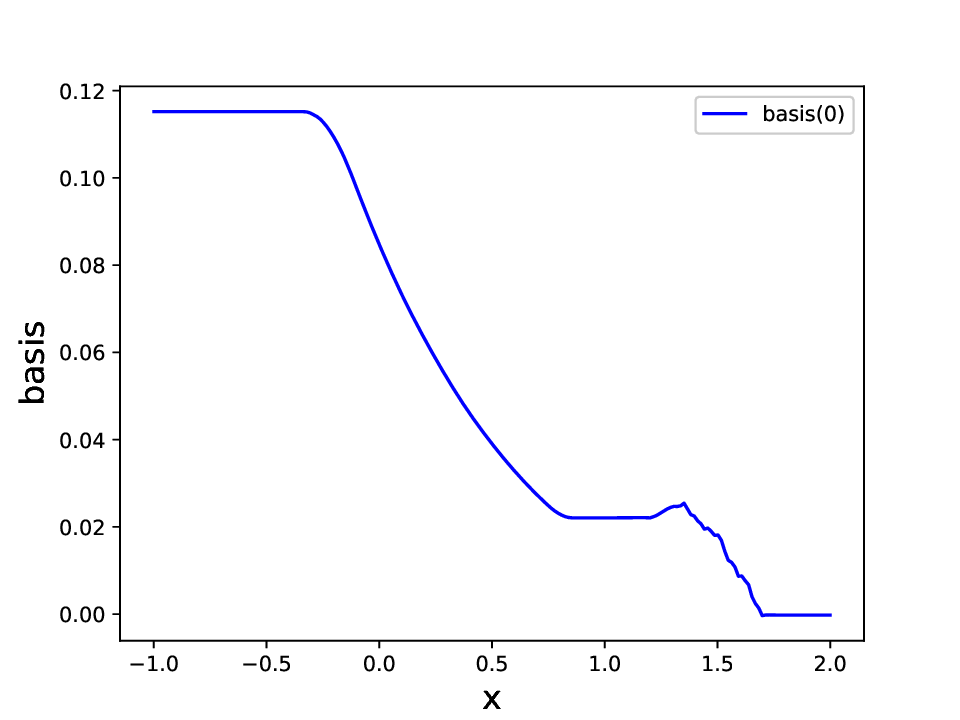}

&
\includegraphics[width=0.33\textwidth,height=0.31\textwidth]{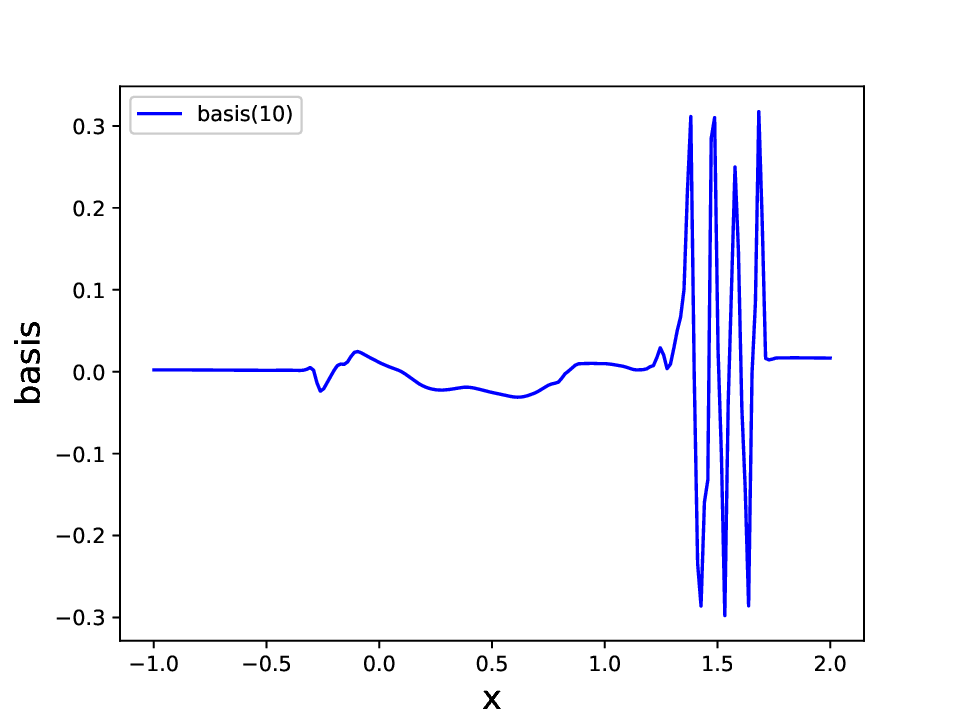}
    
  \\
  (a) Spectrum&  (b) Mode$=0$ basis& (c) Mode$=10$ basis
\\
  \includegraphics[width=0.33\textwidth,height=0.31\textwidth]{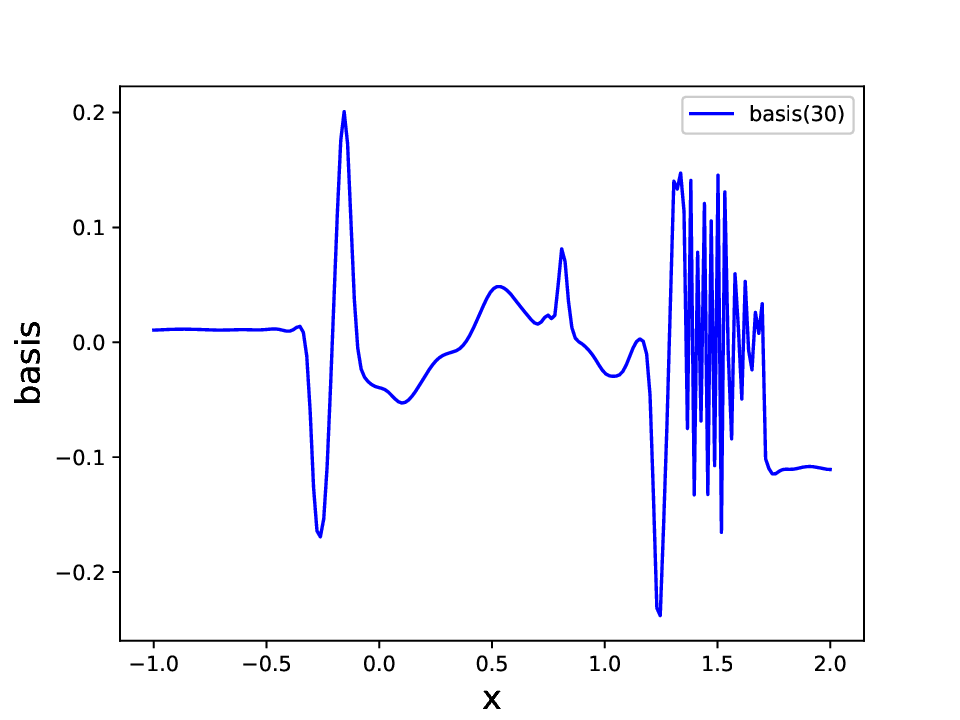}
 &
\includegraphics[width=0.33\textwidth,height=0.31\textwidth]{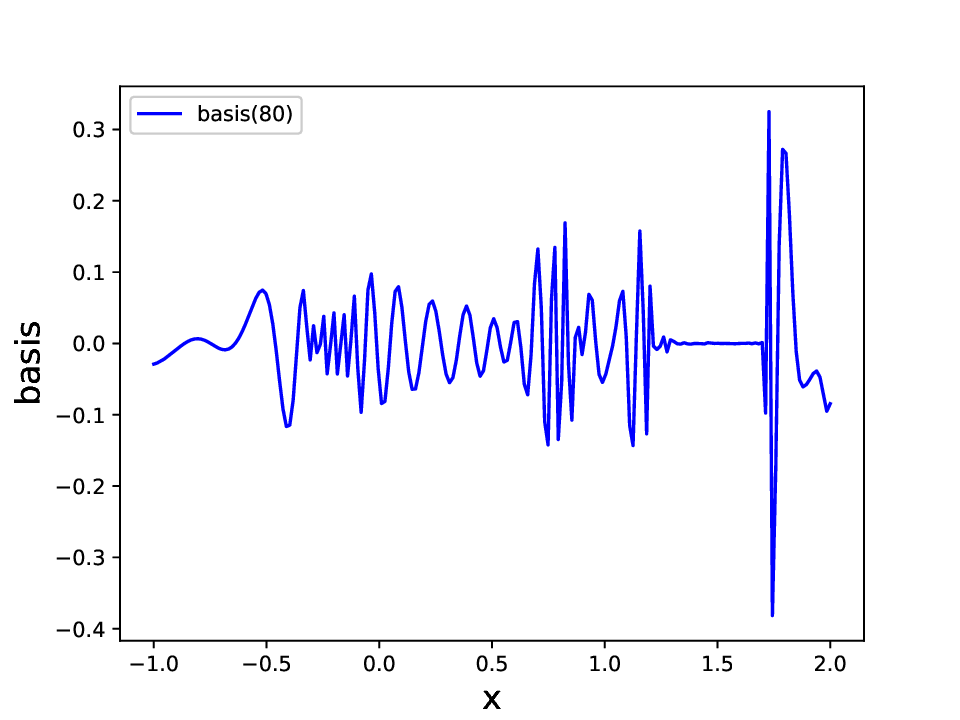}

&
\includegraphics[width=0.33\textwidth,height=0.31\textwidth]{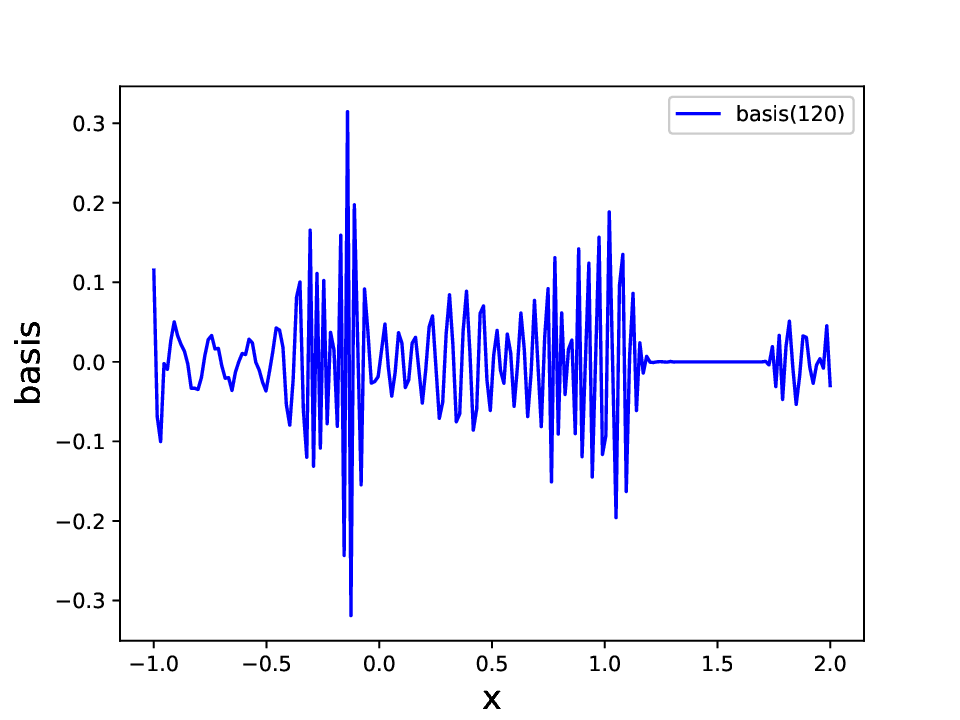}
    
  \\
  (d) Mode$=30$ basis&  (e) Mode$=80$ basis& (f) Mode$=120$ basis

\end{tabular} 

\caption{The top-left plot shows the SVD spectrum for the intermediate pressure ratio, and rest of the plots shows different SVD basis functions.}
    
    \label{fig:hierarchy_inter}
  \end{center}
  
\end{figure}

In Fig.~\ref{fig:hierarchy_inter} we present the eigenvalue spectrum of the basis functions for the IPR problem across 150 modes corresponding to the number of neurons in the last layer of the trunk net. According to Fig.~\ref{fig:hierarchy_inter}(a), the eigenvalues decay when the mode number increases. The descending trend of the spectrum describes the shapes of the basis functions for modes $0$, $10$, $30$, $80$, and $120$. From mode $0$, it is evident that the expansion waves are the first feature that are extracted from the data set. The highest-mode basis functions capture the wide range of frequencies exhibited at discontinuities such as contact and shock waves. Considering Fig.~\ref{fig:hierarchy_inter}(c)-(e), we can realize the fact that these modes are hierarchically contributing to resolving the shock wave feature. The order of the SVD orthonormal basis functions can be employed to construct the inferred solution manually using a specific number of high modes to avoid oscillations at the discontinuities of the solution. In the following section, we explain an approach that modifies the number of basis functions to infer the solution accurately.

\begin{figure}[!t]
  \begin{center}
    \begin{tabular}{ccc}   \includegraphics[width=0.33\textwidth,height=0.31\textwidth]{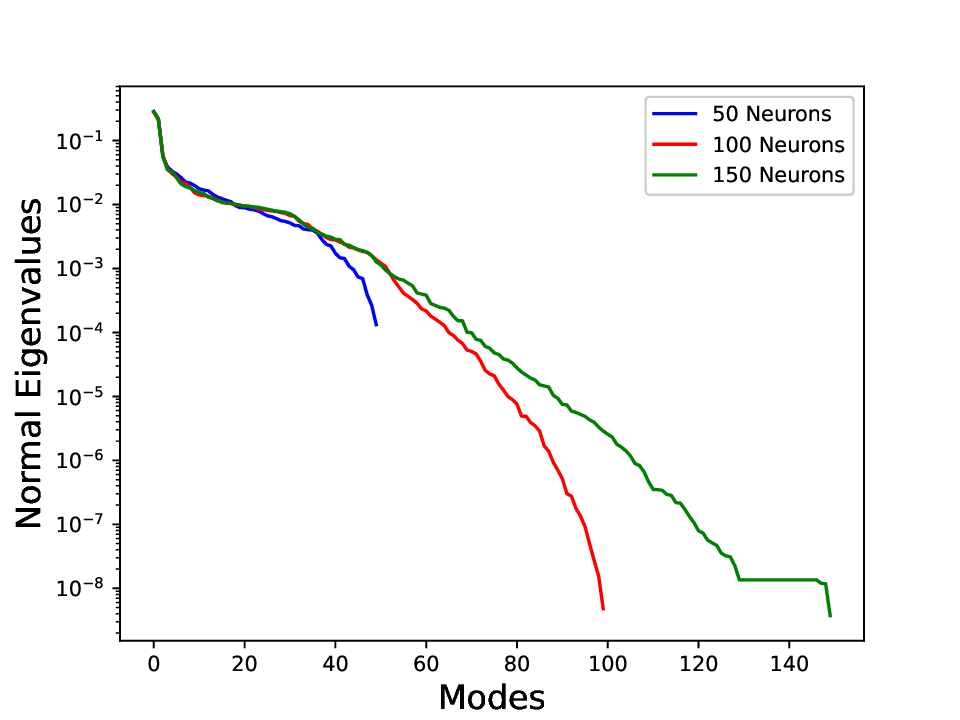}
 &
\includegraphics[width=0.33\textwidth,height=0.31\textwidth]{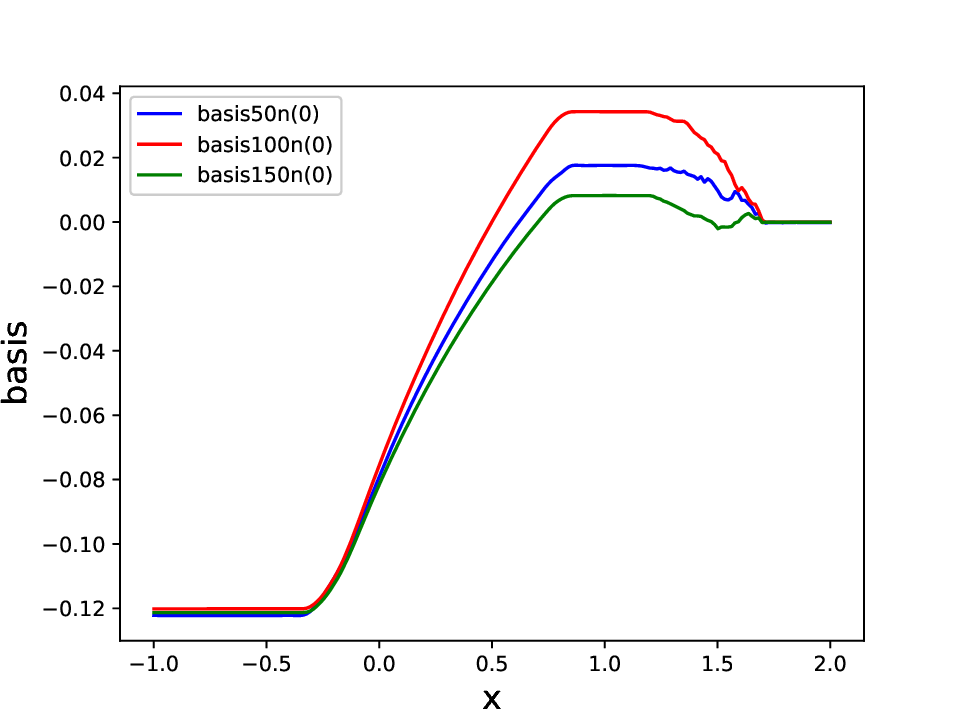}

&
\includegraphics[width=0.33\textwidth,height=0.31\textwidth]{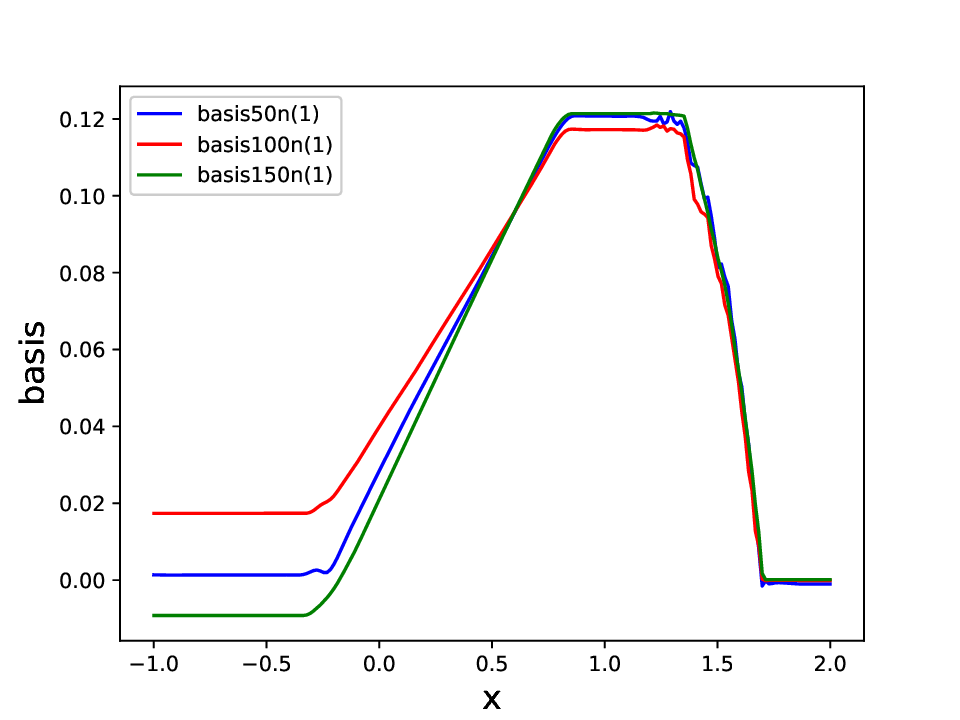}
    
  \\
  (a) Spectrum&  (b) Mode$=0$ basis& (c) Mode$=1$ basis
\\
  \includegraphics[width=0.33\textwidth,height=0.31\textwidth]{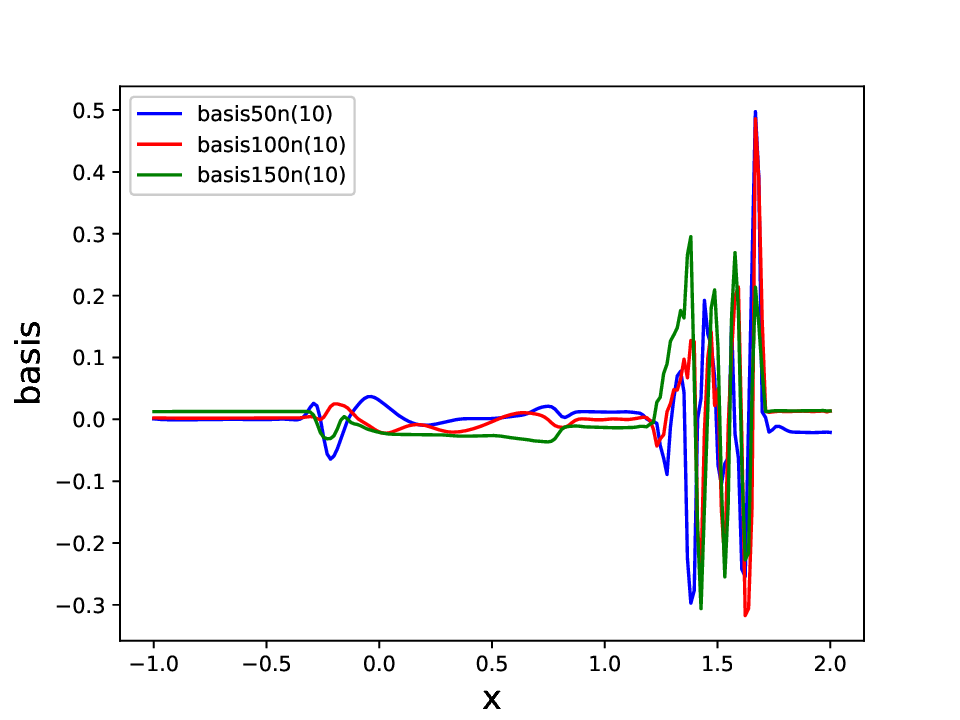}
 &
\includegraphics[width=0.33\textwidth,height=0.31\textwidth]{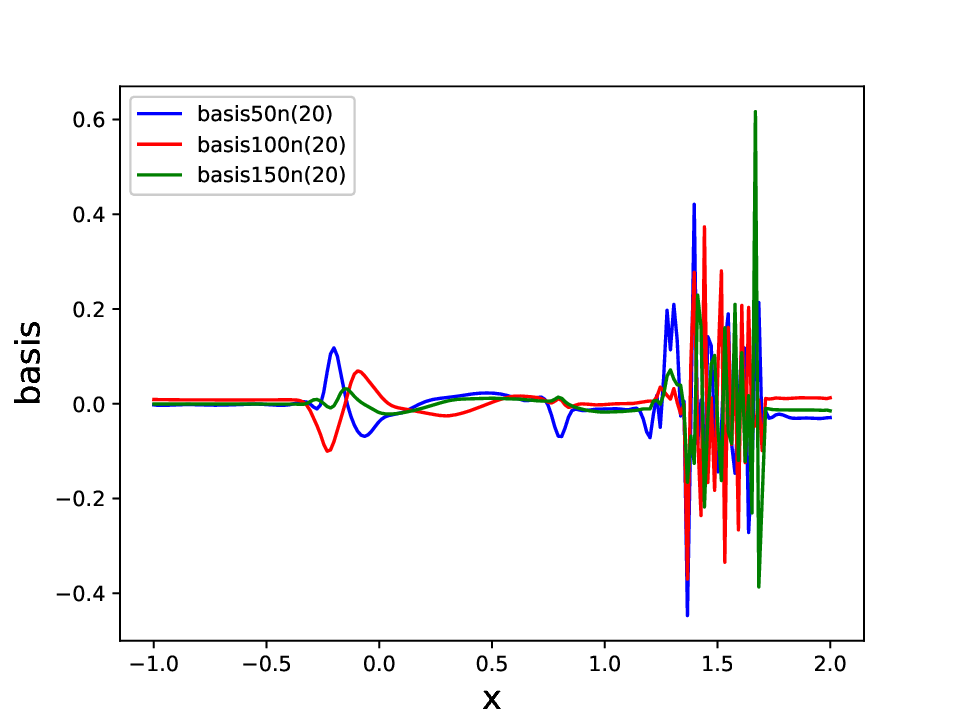}

&
\includegraphics[width=0.33\textwidth,height=0.31\textwidth]{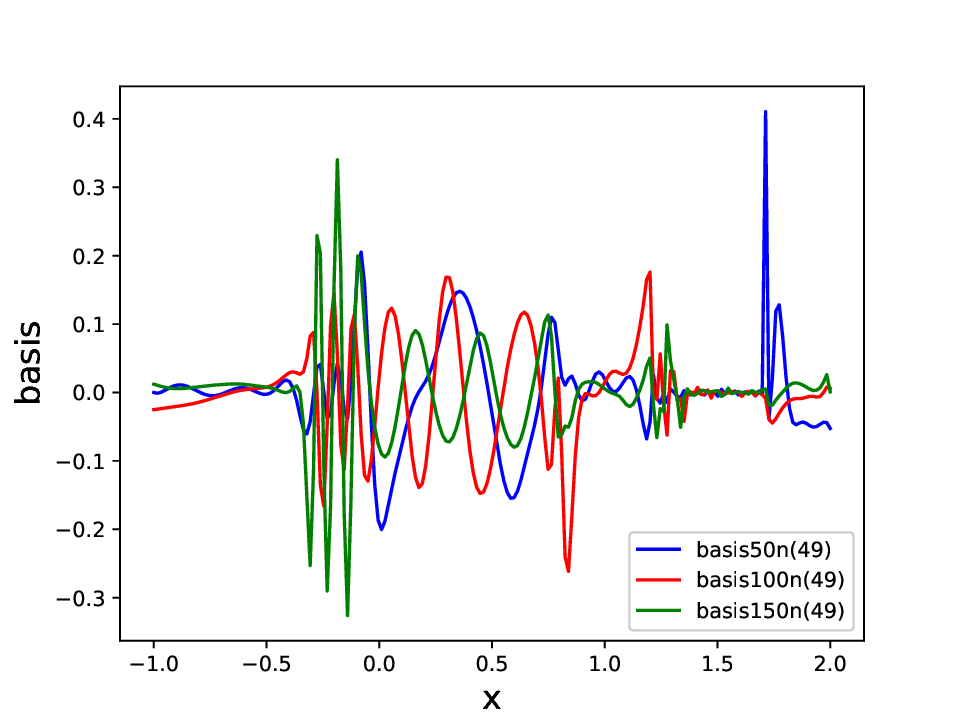}
    
  \\
  (d) Mode$=10$ basis&  (e) Mode$=20$ basis& (f) Mode$=49$ basis

\end{tabular} 

\caption{ Comparison of the SVD spectrum and the basis functions for different number of neurons. It is evident that as the number of neurons increases, more high frequencies are captured, which is useful for accurately resolving shock structure present in the solution.}
    
    \label{fig:numneurons}
  \end{center}
  
\end{figure}

Fig.~\ref{fig:numneurons}(a)-(f) depicts the spectrum of eigenvalues for a trunk net constructed by five hidden layers with a constant width of 50, 100, and 150 neurons corresponding to 50, 100, and 150 eigenvalues and basis functions. By comparing the spectrum of eigenvalues, we observe that using a higher number of modes can recover a higher content of information at the higher modes.  Employing a higher number of neurons in the trunk net layers decreases the $L_2$ norm of the error, but it could induce oscillations from the highest modes. According to Fig.~\ref{fig:numneurons}(b) and (c), the first feature extracted from the data is the expansion wave, while the second one is the shape of the velocity profile. Considering Fig.~\ref{fig:numneurons}(e) and (f), we can deduce that by increasing the number of neurons, we are increasing the capacity of capturing the shock wave more accurately since we can observe high-frequency jumps in the basis function of 50 neuron cases close to the location of the shock wave near the right boundary exits for Mode$=49$. At the same time, there is no high-frequency oscillation for the 100 and 150-neuron cases. The shock wave feature is captured using modes higher than $49$ in the 100 and 150-neuron cases.

\begin{figure}[!t]
  \begin{center}
    \begin{tabular}{cc}   \includegraphics[width=0.40\textwidth,height=0.38\textwidth]{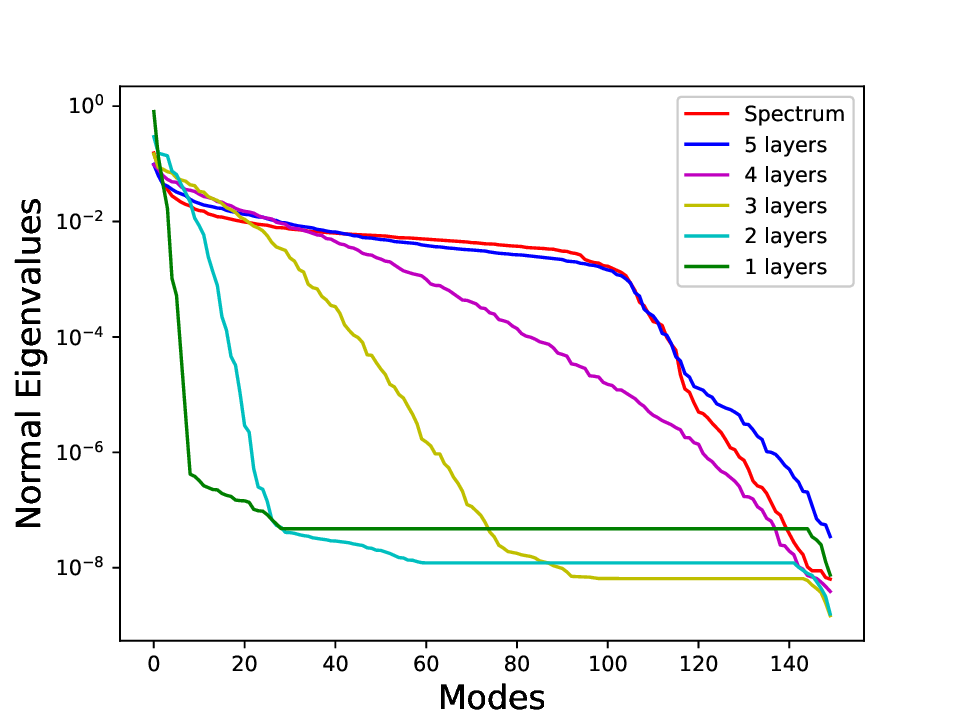}
 &
\includegraphics[width=0.40\textwidth,height=0.38\textwidth]{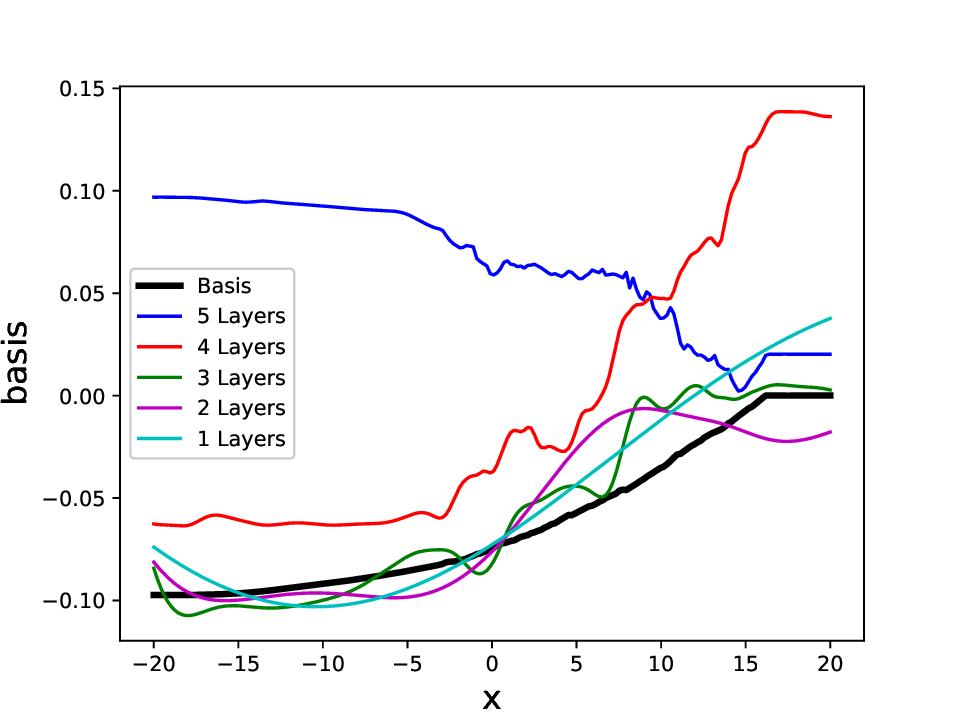}
    
  \\
  (a) Spectrum&  (b) Mode$=0$ basis
\\
  \includegraphics[width=0.40\textwidth,height=0.38\textwidth]{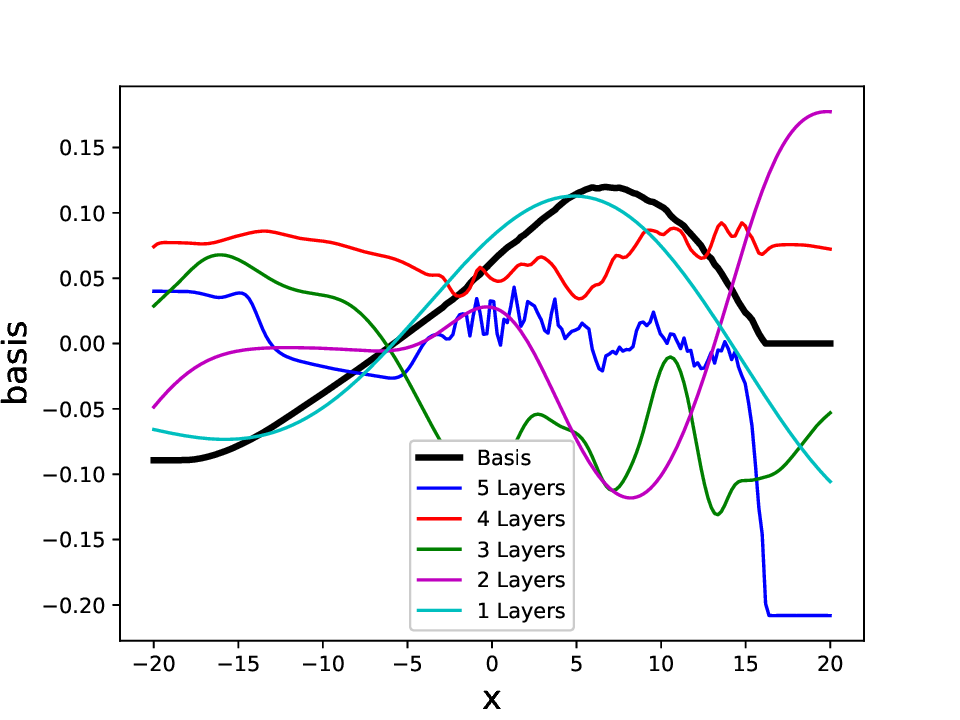}
 &
\includegraphics[width=0.40\textwidth,height=0.38\textwidth]{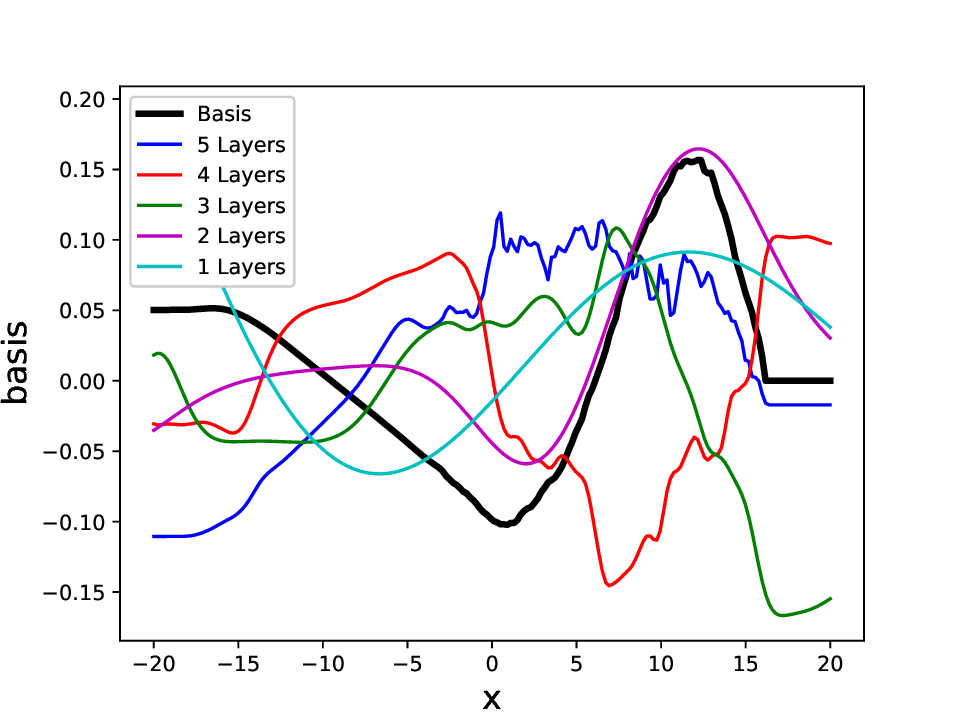}
  \\
  (c) Mode$=1$ basis&  (d) Mode$=2$ basis

\end{tabular} 

\caption{Layer-wise contribution to the spectrum and the basis: The figure shows the contribution of each hidden layer to the final basis. We have plotted the contribution of the first five layers. We see that with each layer, more high-frequencies are captured, and the final layer (layer 5) output (excluding the linear layer) is almost similar to the basis function.}
    
    \label{fig:architecture}
  \end{center}
  
\end{figure}

In Fig.~\ref{fig:architecture}, we attempt to reveal the contribution of each hidden layer to the shape of the final basis functions of the HPR problem. For this purpose, we performed an SVD decomposition of the output of hidden layers from the first to the last linear layer. Using the eigenvalues of each layer output, we can observe the flow of information learned by the trunk net. Considering Fig.~\ref{fig:architecture}(a), the red line shows the spectrum of eigenvalues of the output of the last linear layer of the trunk net. The label ``$m$ layers" refers to the output of the $m^{th}$ hidden layer. The spectrum shows that the first layers always contribute to learning the lowest-mode basis functions. According to Fig.~\ref{fig:architecture}(a), the flow of information does not reach the highest modes until the $4^{th}$ hidden layer, where we can see there is no plateau region in the spectrum of the output of the first $4$, first $5$ and last linear layer. In Fig.~\ref{fig:architecture}(b)-(d), the shape of the first three modes out of the layers is shown. From Fig.~\ref{fig:architecture}(b), we see that the hidden layers keep adding more information as we move from the trunk input to the trunk net output.

\begin{figure}[!t]
  \begin{center}
    \begin{tabular}{cc}   \includegraphics[width=0.35\textwidth,height=0.32\textwidth]{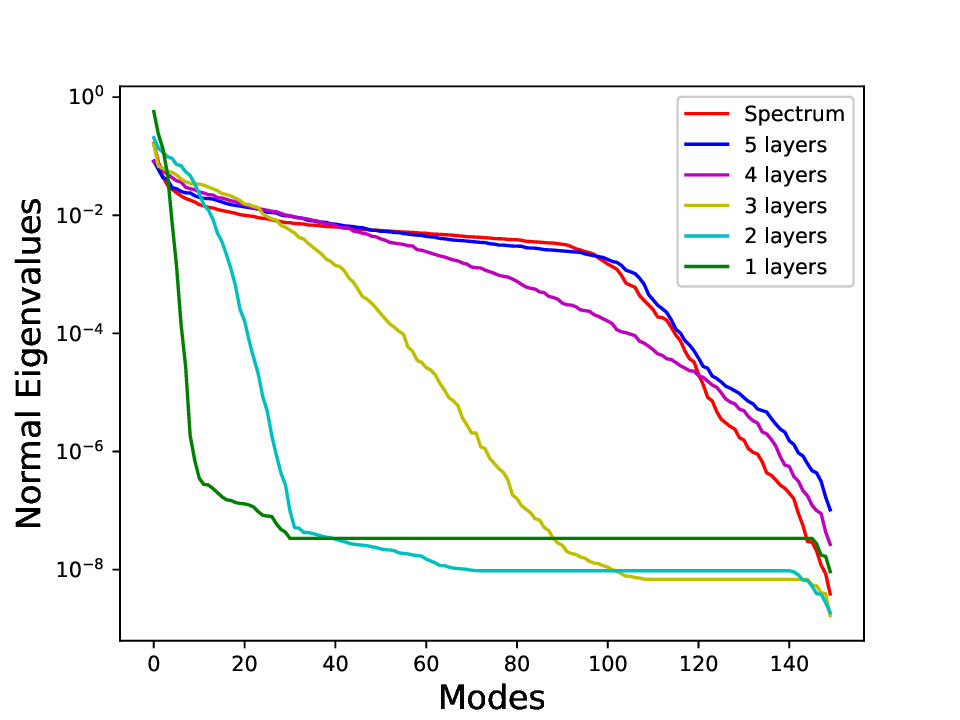}
 &
\includegraphics[width=0.35\textwidth,height=0.32\textwidth]{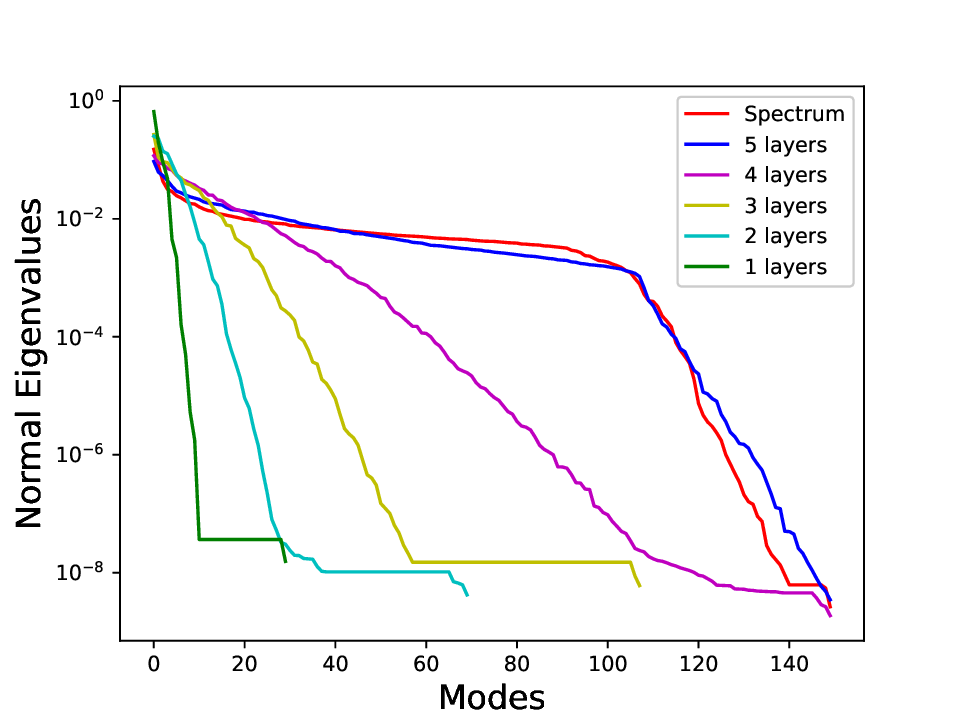}
    
  \\
  (a) Spectrum rectangle architecture&  (b) Spectrum cone architecture
\\
  \includegraphics[width=0.35\textwidth,height=0.32\textwidth]{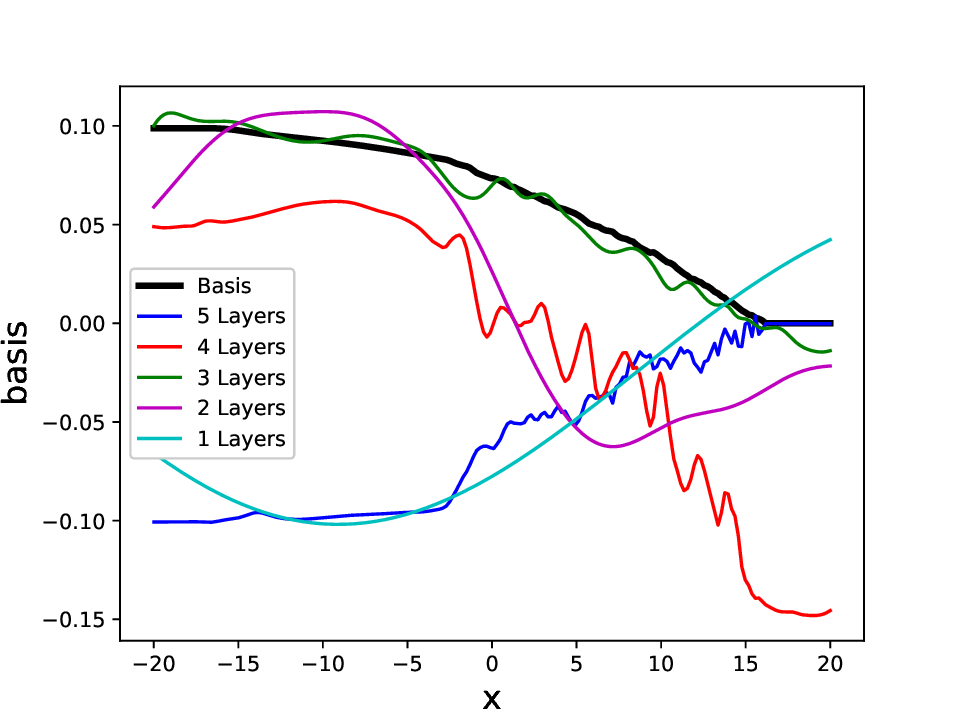}
 &
\includegraphics[width=0.35\textwidth,height=0.32\textwidth]{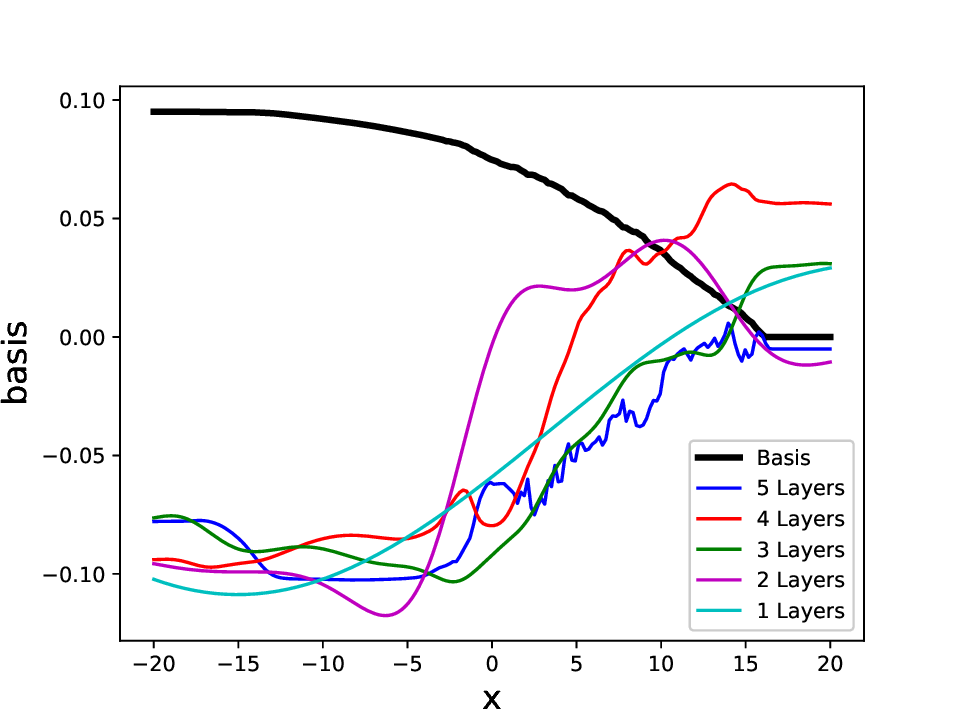}
  \\
  (c) Mode$=0$ basis rectangle architecture &  (d) Mode$=0$ basis cone architecture



\end{tabular} 

\caption{Layer-wise contribution to the spectrum and the basis for rectangular and decoder architectures: The decoder architecture refers to a trunk shape such as [30, 70, 108, 150, 150] and rectangle network refers to [150, 150, 150, 150, 150] hidden layers arrangement. }
    
    \label{fig:contrecvscone}
  \end{center}
  
\end{figure}

According to Fig.~\ref{fig:architecture}(a), we can use a smaller number of neurons for the first hidden layer of the neural net as the eigenvalues of most of the neurons are approximately zero. So, we can generate a cone-shaped configuration of the hidden layers according to the non-zero eigenvalues and use the downsized trunk net for the training procedure. To this end, we chose $[30, 70, 108, 150, 150]$ number of neurons for the hidden layers of the trunk net. We computed the optimal number of neurons for each layer based on the points on the spectrum that reach the plateau state. A comparison of the original rectangle shape versus cone shape architectures is depicted in Fig.~\ref{fig:contrecvscone}. We are comparing the rectangle with a cone shape (Fig.~\ref{fig:contrecvscone}(a) and (b); we can see that when we use the cone shape type, the flow of information to the last layers gets delayed for higher modes. This fact can be seen by comparing the shape of the basis functions of the first $5$ hidden layers. We can see that the shape of the basis functions changes for higher modes in the rectangle architecture.

\begin{figure}[!t]
  \begin{center}
    \begin{tabular}{ccc}   \includegraphics[width=0.33\textwidth,height=0.27\textwidth]{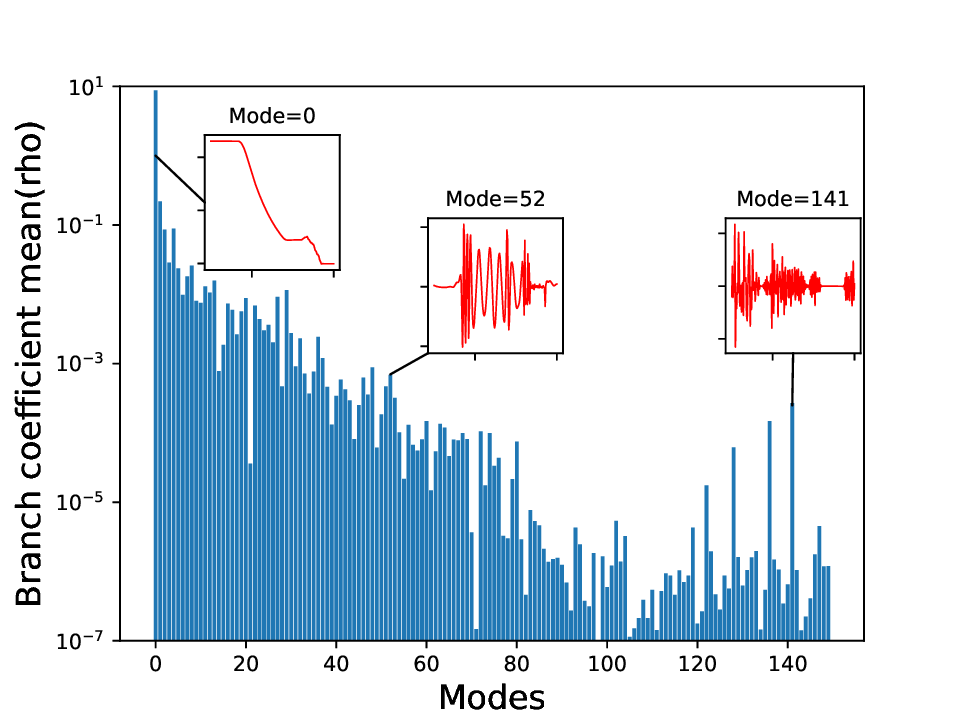}
 &
\includegraphics[width=0.33\textwidth,height=0.27\textwidth]{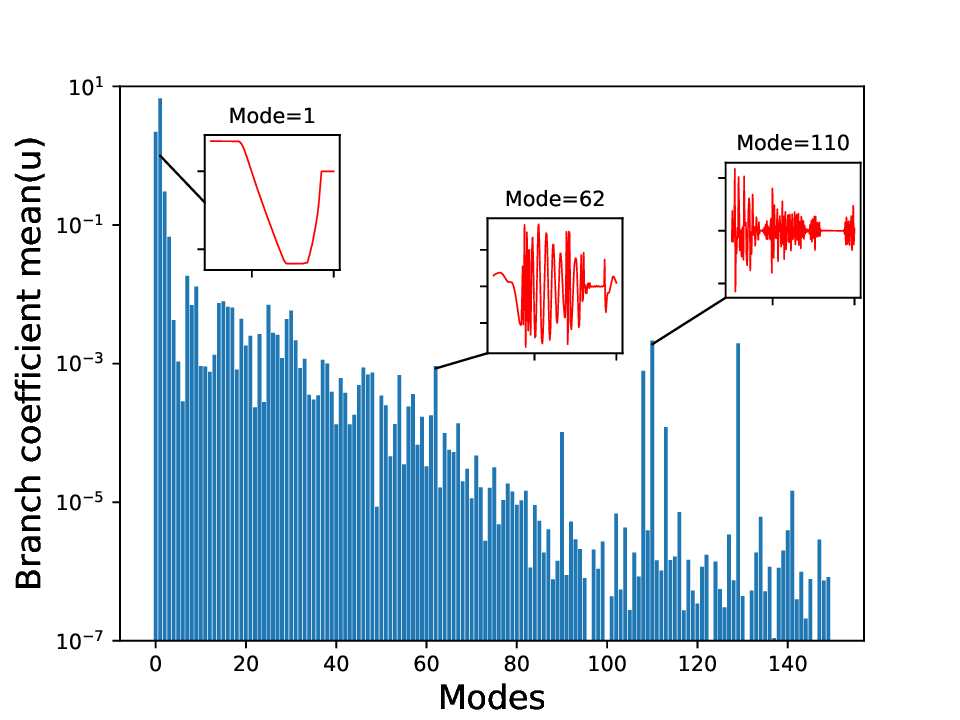}

&
\includegraphics[width=0.33\textwidth,height=0.27\textwidth]{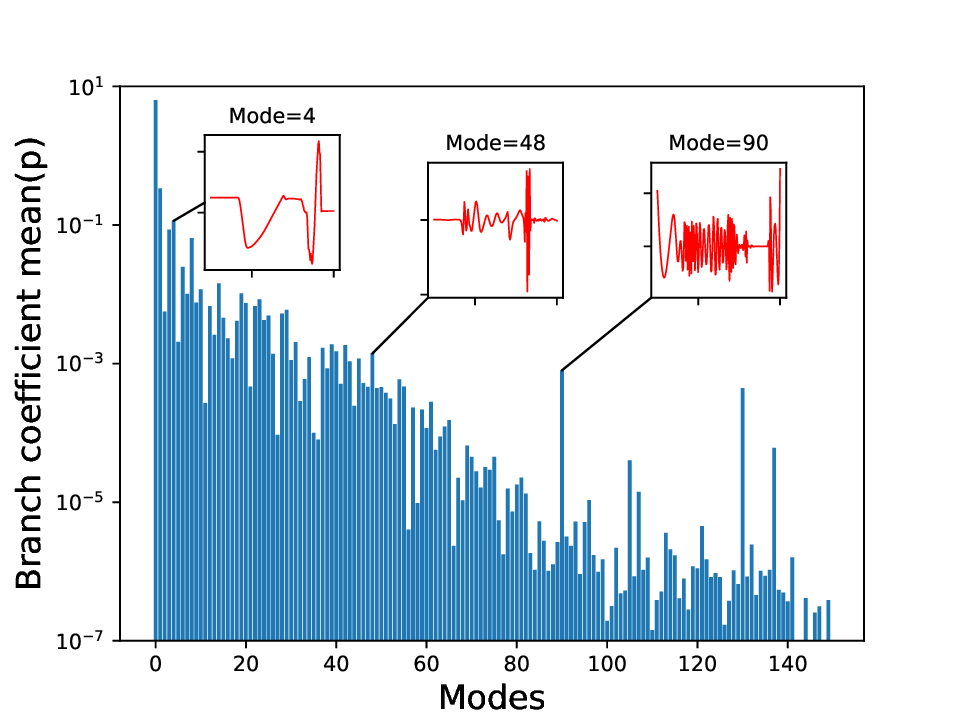}
    
  \\
  (a) Branch coefficients $\rho$ train&  (b) Branch coefficients $u$ train& (c) Branch coefficients $p$ train
\\
  \includegraphics[width=0.33\textwidth,height=0.27\textwidth]{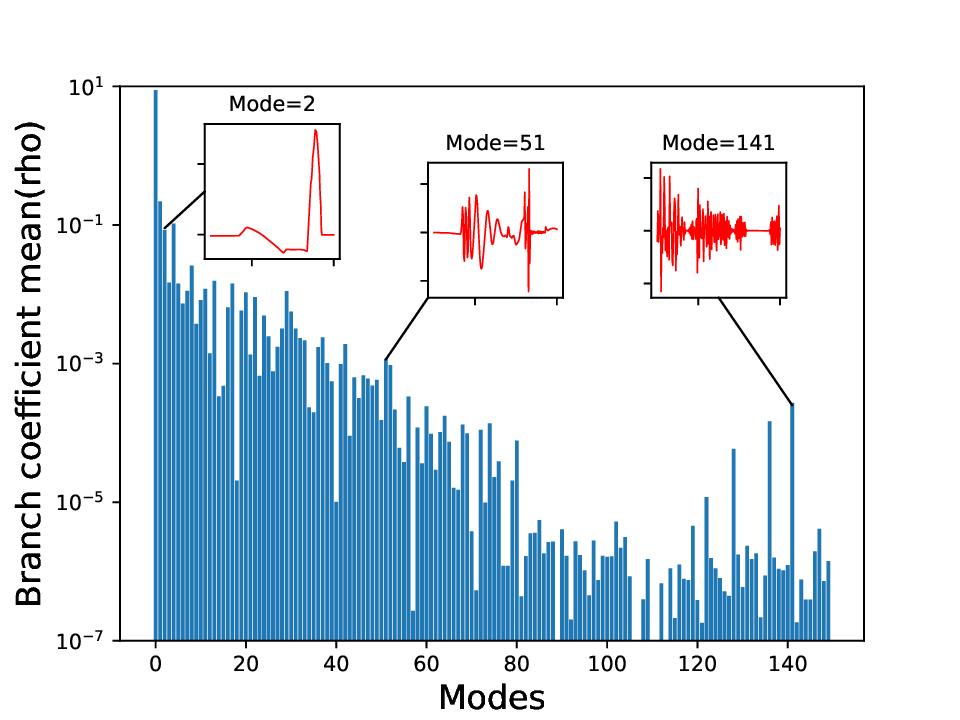}
 &
\includegraphics[width=0.33\textwidth,height=0.27\textwidth]{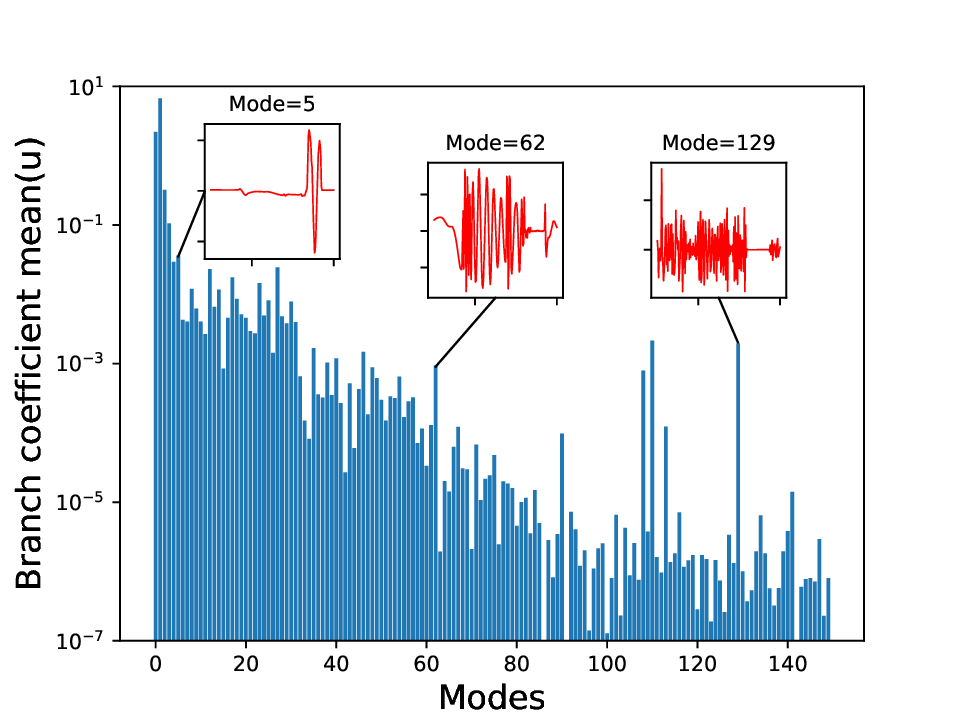}

&
\includegraphics[width=0.33\textwidth,height=0.27\textwidth]{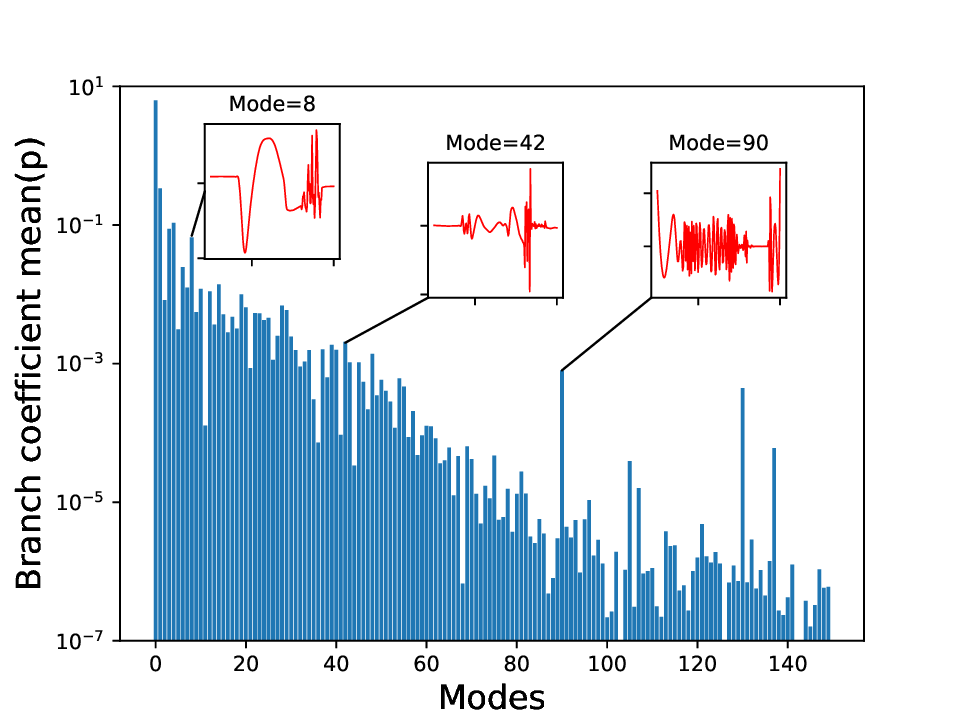}
    
  \\
   (d) Branch coefficients $\rho$ test&  (e) Branch coefficients $u$ test& (f) Branch coefficients $p$ test

\end{tabular} 

\caption{Contribution of coefficients of basis functions in the final inferred solution of the IPR problem: We plotted the value of branch coefficients for the intermediate pressure ratio case. The top row shows the basis coefficients for training, and the bottom row shows the basis coefficients for testing. The first, second, and third columns give the basis coefficients for density, velocity, and pressure. The insets show representative basis spatial modes.}
    
    \label{fig:histagram}
  \end{center}
  
\end{figure}

Figure~\ref{fig:histagram} depicts the value of the coefficients of the basis functions that are predicted by the branch net. The bar plots show the values on a log scale for the vertical scale. The values are sorted from lowest to highest modes for density, velocity, and pressure. The first row depicts the branch coefficients for the training data set, and the second row shows the branch coefficients for the test data set. The inset plots show the shape of the basis function at a particular mode. Here, we are trying to investigate the contribution of various features to the inferred solution of the IPR problem. According to Fig.\ref{fig:histagram}(a), the highest contribution to the density profile comes from mode $0$ while the highest contribution for the velocity is at mode $1$ (see Fig.~\ref{fig:histagram} (b) and (e)). Another interesting fact is the large contribution of the highest modes in the inferred profile of density, velocity, and pressure. This large contribution results from predicting a discontinuous solution consisting of shock and contact waves.

\begin{figure}[!t]
  \begin{center}
    \begin{tabular}{ccc}   \includegraphics[width=0.33\textwidth,height=0.31\textwidth]{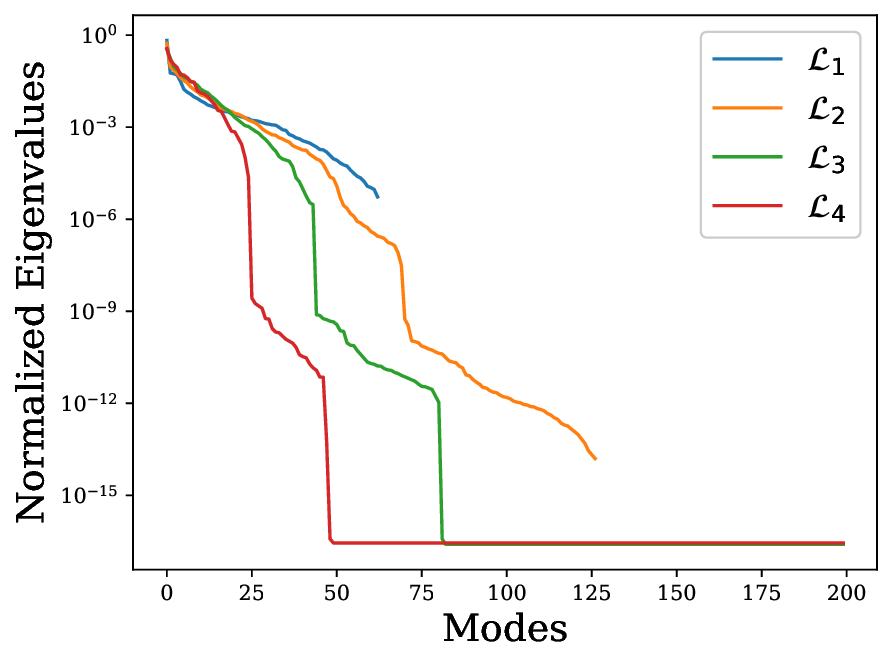}
 &
\includegraphics[width=0.33\textwidth,height=0.31\textwidth]{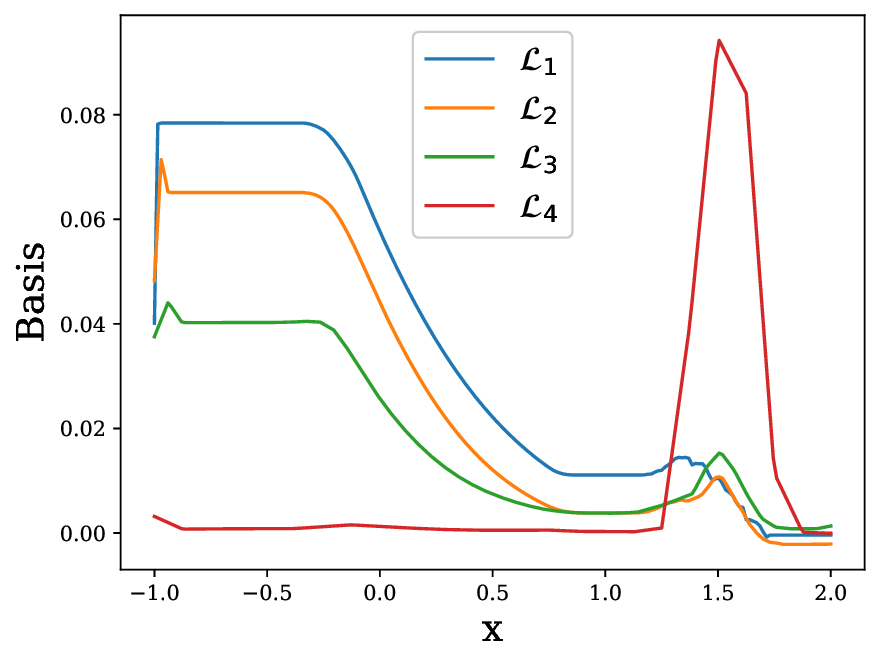}

&
\includegraphics[width=0.33\textwidth,height=0.31\textwidth]{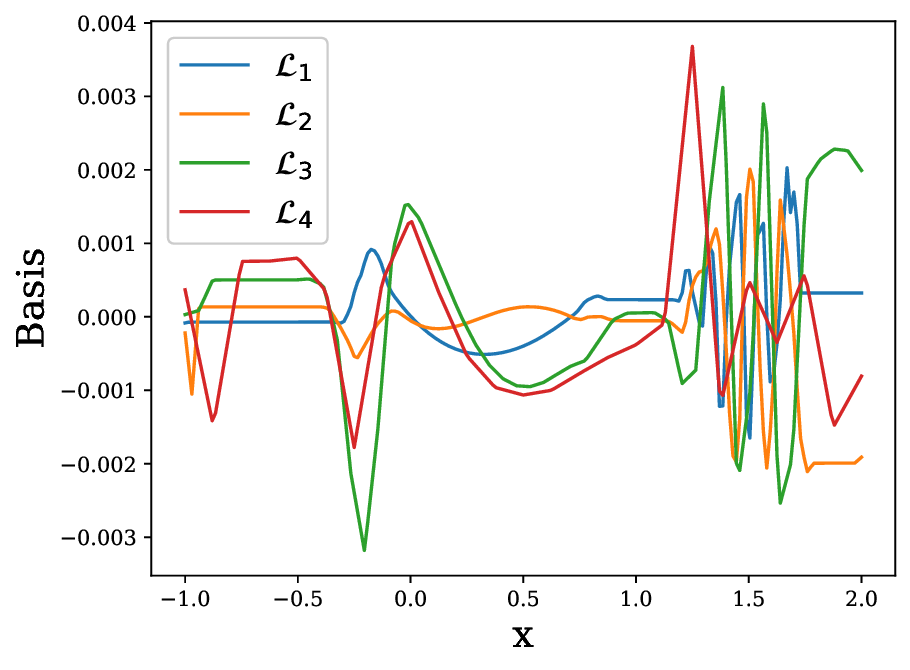}
    
  \\
  (a) Spectrum&  (b) Mode$=0$ basis& (c) Mode$=10$ basis
\\
  \includegraphics[width=0.33\textwidth,height=0.31\textwidth]{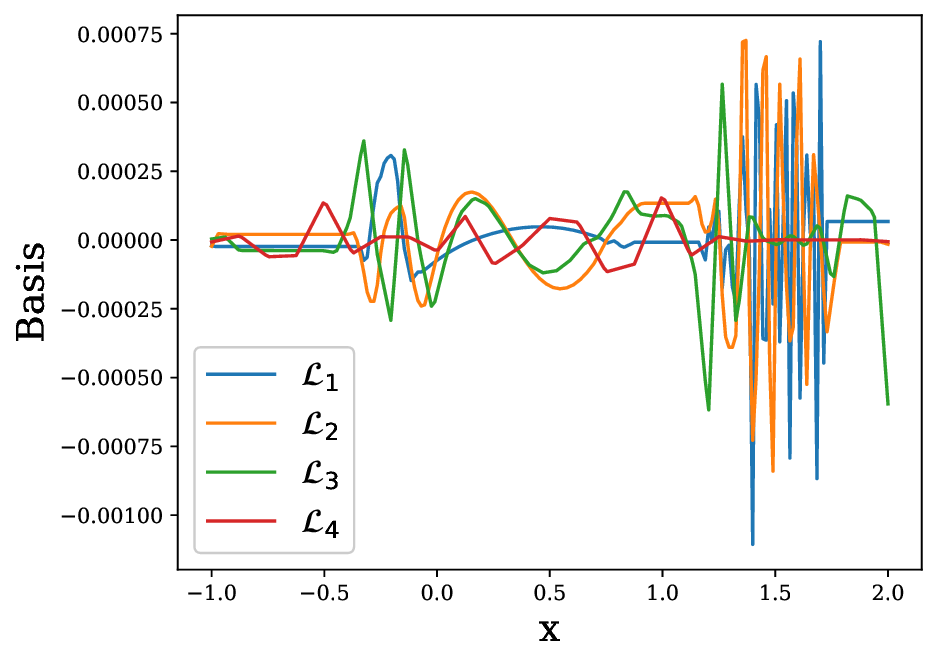}
 &
\includegraphics[width=0.33\textwidth,height=0.31\textwidth]{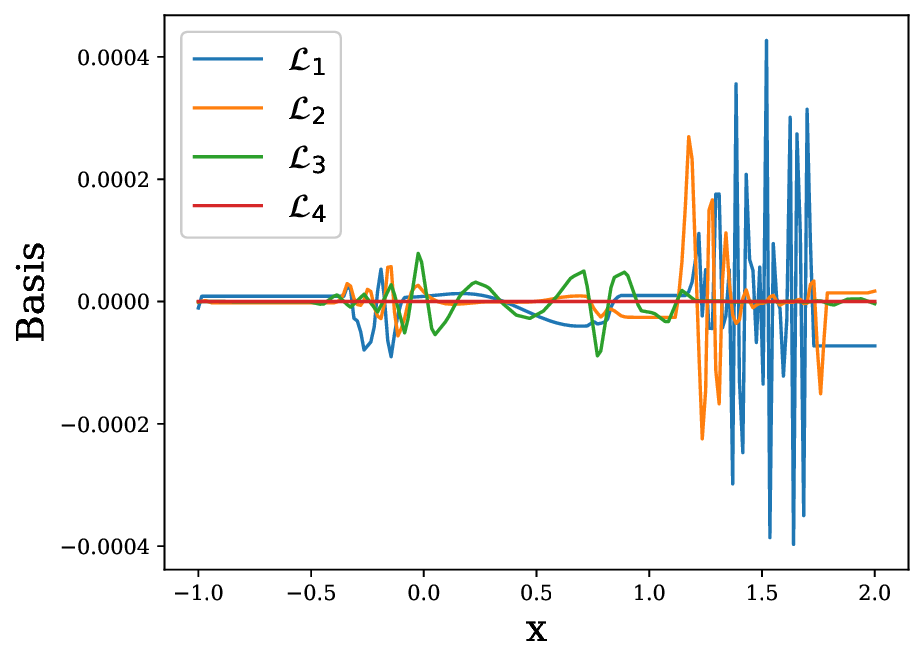}

&
\includegraphics[width=0.33\textwidth,height=0.31\textwidth]{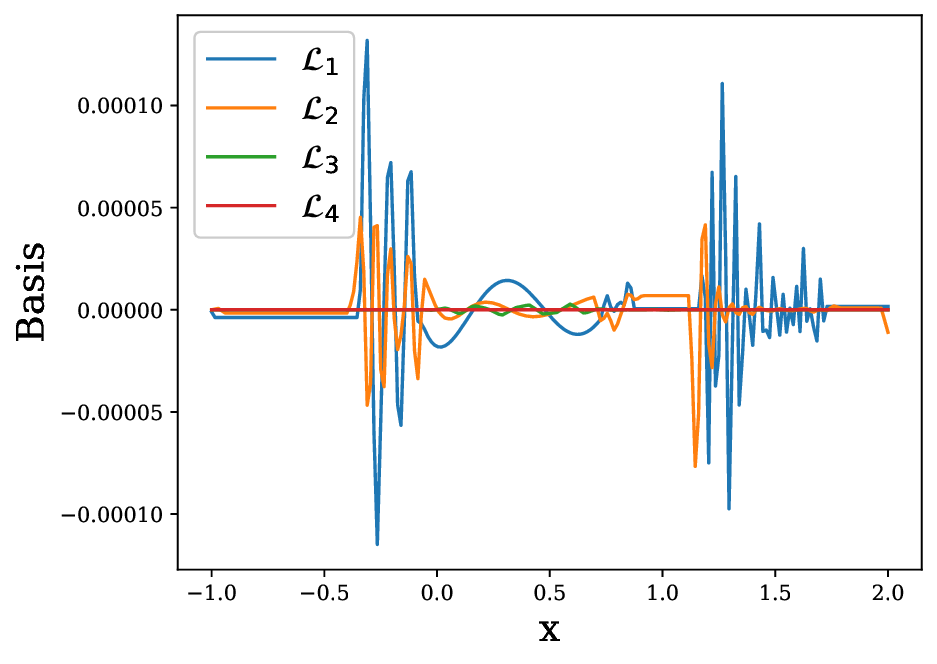}
    
  \\
  (d) Mode$=20$ basis&  (e) Mode$=30$ basis& (f) Mode$=40$ basis

\end{tabular} 

\caption{(a) Eigenspectra of U-Net at $\mathcal{L}_1$, $\mathcal{L}_2$, $\mathcal{L}_3$, $\mathcal{L}_4$. (b)-(f) Orthogonally decomposed representation of the basis learned by the U-Net at different latent levels.}

    \label{fig:unet_svd}
  \end{center}
  
\end{figure}

\subsection{Basis functions - U-Net}
\label{subsec:basis_unet}
In this section, we visualize and analyze the basis functions learned by the U-Net conditioned on $p_l$. 
Specifically, we consider the IPR test case and train the U-Net to learn the dependence of the density field ($\rho$) on $p_l$. 
From Eq.~\eqref{eq:unet_basis}, $\vec{z}^{\mathcal{L}_p}$ represents the discrete basis functions learned by the U-Net at the $p^{th}$ latent level spanning $\mathcal{L}_p$. 
We perform SVD on $\vec{z}^{\mathcal{L}_p}$ separately for $p=1,2,3,4$ and visualize the eigenvalues and eigenmodes.

\begin{equation}
    \vec{z}^{\mathcal{L}_p} = U_p \Sigma_p V_p^T
\end{equation}
where $\Sigma_p$ represents the ordered list of eigenvalues and $V_p$ represents the corresponding eigenmodes at the latent space $\mathcal{L}_p$. 
We visualize the decay of eigen values $\Sigma_p$ for $p=1,2,3,4$ in Fig.~\ref{fig:unet_svd} (a). 
We also plot the scaled basis functions for eigen modes = 0, 10, 20, 30, 40 for all $p$ in Fig.~\ref{fig:unet_svd} (b)-(f).

In Fig.~\ref{fig:unet_svd}(a) we observe that the rate of decay of eigenvalues is greater for $\mathcal{L}_4$ than $\mathcal{L}_3$ than $\mathcal{L}_2$ than $\mathcal{L}_1$. 
Therefore, the basis functions $\vec{z}^{\mathcal{L}_p}$ with a larger $p$ only learn the high energy modes with lower frequencies, and subsequently, the basis functions with a lower $p$ become increasingly responsible for learning the lower energy modes that carry higher frequencies. 
Furthermore, from Fig.~\ref{fig:unet_svd} (e) and (f), the basis functions corresponding to $\mathcal{L}_4$ are constant, which again indicates its inability to capture high frequencies.
The 0$^{th}$ modes of $\mathcal{L}_1$ and $\mathcal{L}_2$ shown in Fig.~\ref{fig:unet_svd} (b) have a striking similarity with the pattern of the density fields learned by the U-Net, as well as the 0$^{th}$ mode learned by DeepONet shown in Fig.~\ref{fig:hierarchy_inter}(b).

\subsection{Constructing optimal sets of basis functions for higher  accuracy}
We investigate the effect of the trunk net width on the accuracy of the two-step approach to learning the solution to the HPR problem. According to Table~\ref{tbl:neurons},  we can conclude that increasing the number of neurons in the trunk network can lead to higher accuracy. Figure ~\ref{fig:numneurons} shows that the accuracy improvement is achieved by incorporating higher modes basis functions to construct the discontinuous solution of the Riemann problems. However, by increasing the number of neurons, we are adding high oscillatory basis functions to the set, which can give rise to unwanted oscillations in the inferred solution. Therefore, at the inference stage, we can take advantage of the hierarchical features of the SVD basis functions and remove higher modes one by one from the basis set. We can monitor the corresponding accuracy by computing the $L_2$ norm of the solution predicted by the truncated set of basis functions. We then use the set of basis functions that provides the lowest $L_2$ norm of the error. In brief, we can design a procedure to use the two-step DeepONet for the best-inferred solution of the Riemann problems as follows:

\begin{table}[!h]
\footnotesize
\caption{Relative L2 norms mean and standard deviation obtained using 10 runs. The $L_2$ norm of the error is calculated over the entire testing dataset for density, velocity, and pressure profiles for the HPR test cases using 50, 100, and 150 neurons for the width of the trunk net. The time reported is the training time; the inference time is negligible.}  
\centering
\begin{tabular}{lccccc} 
\hline
Cases &$L_2(\rho)$ \%&$L_2(u)$ \%&$L_2(p)$ \%& total $L_2$ norm \%& Time (Min)\\
\hline
HPR(2 step Rowdy(50)) & $2.96\pm0.050$   & $6.51\pm0.046$ & $8.47\pm0.147$& $5.98\pm0.080$& $33.23$ \\
HPR(2 step Rowdy(100)) & $0.92\pm0.043$   & $3.83\pm0.147$ & $3.28\pm0.839$& $2.67\pm0.343$& $34.51$ \\
HPR(2 step Rowdy(150)) & $\mathbf{0.66\pm0.093}$   & $\mathbf{3.39\pm0.104}$ & $\mathbf{2.86\pm1.680}$& $\mathbf{2.31\pm0.626}$& $34.64$ \\
\hline
\end{tabular}
\label{tbl:neurons}
\end{table}

\begin{enumerate}
    \item Increase the number of neurons at the last layer of the trunk net to obtain the lowest value of loss during training.
    \item If oscillations are near the discontinuous solution, we remove the highest basis functions from the set and compute the optimal error value.
    \item Use the optimum number of basis functions for the trained model for future inference.
\end{enumerate}

\begin{figure}[!h]
  \begin{center}
\includegraphics[width=0.5\textwidth,height=0.45\textwidth]{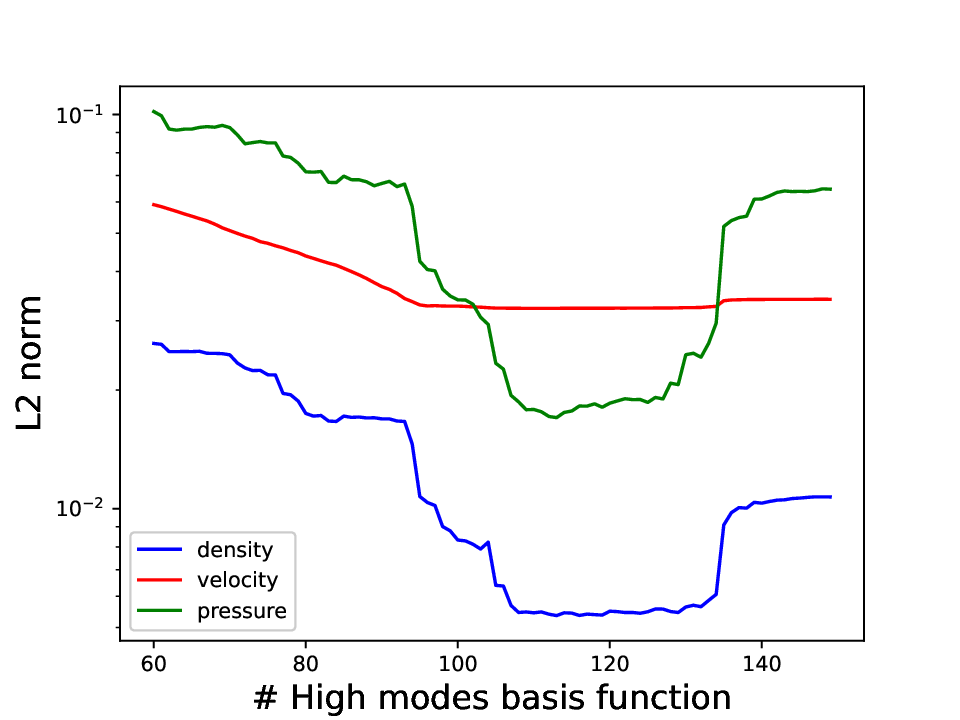}
\caption{ The $L_2$ norm of the error in density, velocity, and pressure for the HPR problem by using optimal number of basis functions to avoid numerical artifact due to the existence of high mode basis functions in the orthonormal set. }
    \label{fig:l2norm}
  \end{center}

\end{figure}
We have followed this procedure on the HPR test case with ten ensembles to ascertain our claim. First, we took the best-trained DeepONet from the set of 10 ensembles, removed the highest mode basis functions from the set, and used the remaining basis functions to infer the solution. We then computed the $L_2$ norm of the variables using the first 60 to 150 basis functions from the orthonormal basis set. The results of the errors are depicted in Fig.~\ref{fig:l2norm}. We can observe that the optimal number of basis functions to use here is 114 first basis functions. We applied the approach to the ten ensembles of the two step-trained DeepONets and recalculated the $\%L_2(\rho)=0.573\pm0.091$ ,$\%L_2(u)=3.39\pm0.105$, and $\%L_2(p)=2.29\pm1.78$. We improved the accuracy of $\rho$ by $13.6$ percent and of $p$ by $20.0$ percent. The accuracy of the velocity remained intact, as we may need to use an even higher number than 150 neurons to see a similar trend as in pressure and density inference results. 



\section{Summary}
Enhancing the prediction of solutions for high-speed flows governed by compressible Euler equations holds significant implications for the design of aerospace vehicles, including airplanes, re-entry vehicles, missiles, and more. In this study, we leveraged the properties of deep neural operators -- RiemannONets -- to tackle Riemann problems, crucial for simulating high-speed flows. These problems entail discontinuous solutions, such as shocks and contact discontinuities, representing some of the most challenging aspects in the realm of scientific computing.

We devised the RiemannONets by incorporating two distinct neural operators. The first operator is built upon DeepONet, which underwent modifications for a two-stage training approach, enhancing prediction accuracy. The second operator is a U-Net, tailored to be conditioned on pressure initialization. Our training and testing of RiemannONets focused on the Sod shock tube problem, encompassing pressure ratios ranging from 10 to $10^{10}$. Specifically, RiemannONets were trained on input-output data sets, establishing a mapping from the initial solution to the final time step. Once trained, RiemannONets can predict solutions for unseen datasets in real-time, requiring no additional optimization.
The predictions from both neural operators exhibited remarkable accuracy across low, intermediate, and very high pressure ratios, with an error margin below $2\%$. However, it is noteworthy that U-Net's computational speed lags behind DeepONet by orders of magnitude. These results correspond to a single mapping from the initial condition to the final time. We have also shown that the use of adaptive activation function in the structure of the DeepONet increases the accuracy of the prediction comparing to the fixed activation functions. Notably, we achieved similarly accurate results for the time-evolving Sod problem using the DeepONet trained in two stages.

We systematically explored the basis functions generated by the trunk net of DeepONet for representing the operator in a continuous manner. Using the orthonormalization process, we constructed a set of orthonormal basis functions. For orthonormalization, we performed QR-factorization and  singular value decomposition (SVD) on the output of trunk net. Our in-depth   investigation of the data-driven basis functions has led us to the following conclusions:

\begin{itemize}
    \item The SVD decomposition resulted in a hierarchical orthonormal basis with distinctive features, which can be used to remove Gibbs oscillations at the discontinuities during the inference.
    \item A comparison of the eigenvalue spectra of QR and SVD revealed that the QR eigenvalues are equally contributing to the final solution, while the SVD eigenvalues show a descending trend in contribution from low to high modes.
    \item The unique basis of the SVD decomposition proves particularly advantageous for high-speed flows.
    \item The SVD basis functions exhibit a similar shape for low, intermediate, and high-pressure ratio Sod problems. Consequently, we can utilize a trunk network trained on a fixed range of pressure ratios to train the branch network for higher pressure ratios ranges.
    \item Using a larger width of the trunk network in the two-step training improves the accuracy of the inference by learning a higher amount of information at high modes.
    \item The first hidden layers of the trunk network are responsible for learning low modes while the later hidden layers contribute to learning the high-frequency features.
    \item The specific contribution of each basis function in constructing density, velocity, and pressure fields are explored, revealing the distinct hierarchy of feature learning for density, velocity, and pressure profiles.
    \item Employing the hierarchy of the SVD basis functions, we constructed a procedure that can effectively remove high frequency artifacts near the discontinuous region of the Riemann problem solutions. 
\end{itemize}
At present, there are no numerical methods in existence that can attain the SVD data-driven orthonormal basis functions. This requires the incorporation of ad hoc features in the basis functions, such as the utilization of enrichment methods in finite elements or the integration of a feature layer in a neural network to address discontinuous, singular, or multiscale solutions. Additionally, we visualize the basis functions learned by different latent levels of the U-Net. This rvealed that the first mode of U-Net is similar to the first mode of DeepOnet. Moreover, the lower latent layers focus on learning the low modes whereas the higher levels capture the highest modes.

Our ongoing research is focused on examining basis functions tailored for addressing two- and three-dimensional high-speed flow problems characterized by discontinuous solutions. 


\section*{Acknowledgments}
This work was supported by the U.S. Army Research Laboratory
W911NF-22-2-0047 and by the MURI-AFOSR FA9550-20-1-0358. 




\bibliographystyle{elsarticle-num} 
\bibliography{reference}
\end{document}